\documentclass{article}

\usepackage[final,nonatbib]{neurips_2024}

\usepackage[utf8]{inputenc} 
\usepackage[T1]{fontenc}    
\usepackage[colorlinks]{hyperref}       
\usepackage{url}            
\usepackage{booktabs}       
\usepackage{amsfonts}       
\usepackage{nicefrac}       
\usepackage{microtype}      
\usepackage{xcolor}         
\usepackage{algorithm}
\usepackage{algpseudocode}
\usepackage{graphicx}
\usepackage{enumitem}
\usepackage{multirow}
\usepackage{makecell}
\usepackage{subcaption}
\usepackage{tabularx}
\usepackage{amsmath}
\usepackage{amsfonts}
\usepackage{soul}
\usepackage{listings}
\usepackage[labelfont=bf]{caption}
\algrenewcommand\algorithmicensure{\textbf{Input:}}

\usepackage{xcolor} 
\usepackage{tcolorbox} 
\definecolor{pinkobject}{RGB}{227,106,216}
\definecolor{blueobject}{RGB}{128,152,238}
\definecolor{purplefp}{RGB}{162,138,202}
\definecolor{bluefp}{RGB}{128,169,205}

\newcommand{\stext}[1]{\text{\raisebox{0pt}[0pt][0pt]{#1}}}
\newcommand{\sigmamax}{\sigma_{\stext{max}}}
\newcommand{\sigmamin}{\sigma_{\stext{min}}}
\newcommand{\sigmadata}{\sigma_{\stext{data}}}

\newcommand{\Smin}{S_{\stext{min}}}

\newcommand{\xx}{\boldsymbol{x}}
\newcommand{\yy}{\boldsymbol{y}}
\newcommand{\cc}{\boldsymbol{c}}
\newcommand{\pp}{\boldsymbol{p}}
\newcommand{\rr}{\boldsymbol{r}}
\newcommand{\dd}{\boldsymbol{d}}
\newcommand{\bg}{\boldsymbol{g}}
\newcommand{\pdrop}{p_{\stext{drop}}}
\newcommand{\Tsse}{T_{\stext{sse}}}

\title{DeBaRA: Denoising-Based 3D Room Arrangement Generation}

\author{%
  Léopold~Maillard$^{1,2}$ \quad Nicolas~Sereyjol-Garros\thanks{Work done during internship at Dassault Systèmes.} \quad Tom~Durand$^{2}$ \quad Maks~Ovsjanikov$^{1}$\\
  \\
  \textsuperscript{1}LIX, École~Polytechnique, IP~Paris \quad \textsuperscript{2}Dassault~Systèmes \vspace{2pt}
  \\
  \texttt{\small \{maillard,maks\}@lix.polytechnique.fr \quad \{firsname.lastname\}@3ds.com}\\
}

\begin{document}

\maketitle

\begin{abstract}
  Generating realistic and diverse layouts of furnished indoor 3D scenes unlocks multiple interactive applications impacting a wide range of industries. The inherent complexity of object \textit{interactions}, the limited amount of available data and the requirement to fulfill spatial constraints all make generative modeling for 3D scene synthesis and arrangement challenging. Current methods address these challenges autoregressively or by using off-the-shelf diffusion objectives by simultaneously predicting all attributes without 3D reasoning considerations. In this paper, we introduce DeBaRA, a score-based model specifically tailored for precise, controllable and flexible arrangement generation in a bounded environment. We argue that the most critical component of a scene synthesis system is to accurately establish the \textit{size} and \textit{position} of various objects within a restricted area. Based on this insight, we propose a lightweight conditional score-based model designed with 3D spatial awareness at its core. We demonstrate that by focusing on spatial attributes of objects, a single trained DeBaRA model can be leveraged at test time to perform several downstream applications such as scene synthesis, completion and re-arrangement. Further, we introduce a novel \textit{Self Score Evaluation} procedure so it can be optimally employed alongside external LLM models. We evaluate our approach through extensive experiments and demonstrate significant improvement upon state-of-the-art approaches in a range of scenarios.
\end{abstract}

\section{Introduction}

Systems capable of generating realistic environments comprising multiple interacting objects would impact several industries including video games, robotics, augmented and virtual reality (AR/VR) and computer-aided interior design. As a result and in tandem with the growing availability of synthetic datasets of indoor layouts~\cite{3dfront,hypersim,habitat,furniscene,scannet}, which can be populated with high-quality 3D assets~\cite{3dfuture,furniscene,scan2cad}, data-driven approaches for automatically generating and arranging 3D scenes have been actively investigated by the computer vision community. Notably, the ongoing success of deep generative models for controllable content creation in the text and image domains has recently been extended to scene synthesis, allowing users to craft realistic indoor environments from a set of multimodal constraints~\cite{atiss,cofs,sceneformer,planit,cliplayout,3Dsp2Ds}.

Challenges associated with 3D indoor scene generation are numerous as the intricate nature of multi-object interactions is difficult to capture and model precisely. Items should be placed, potentially resized and oriented relative to one another, in a way that is both plausible and aligned with subjective and context-dependent priors such as style, as well as ergonomic and functional preferences. Additionally, objects should fit within a bounded, restricted area, and a subtle mismatch can break the perceived validity of the synthesized environment (e.g., overlapping, floating or out-of-bounds objects, inaccessible areas). Finally, the limited availability of high-quality data~\cite{3dfront,habitat} requires learning-based approaches to make careful design choices and trade-offs.

Early data-driven approaches often rely on intermediate hand-crafted representations~\cite{fastsynth,planit,wonkagraph,commonscenes} that are closely related to the considered dataset, which introduces significant biases. Concurrently, popular methods have been adopting autoregressive architectures that treat scene synthesis as a set generation task~\cite{sceneformer,atiss,cliplayout,layoutenhancer,cofs} by sequentially adding individual objects. More recently, score-based generative models (also known as \textit{denoising diffusion models}) have shown promising capabilities in various 3D scene understanding applications~\cite{huang2023diffusion,blockfusion} including controllable scene synthesis~\cite{diffuscene,commonscenes,yang2024physcene} and re-arrangement~\cite{legonet}. In contrast to previous methods, denoising-based approaches enable a stable and scalable training phase and can output all scene attributes simultaneously. The  iterative sampling framework brings an improved consideration for the conditioning information and an attractive balance between generation quality and variety. However, current methods leveraging score-based generative models try to model all attributes (both categorical and spatial) within a single framework, which, as we demonstrate below, is less data-efficient and leads to suboptimal solutions.

In this context, our work aims to establish principled and robust capabilities for generating accurate and diverse 3D layouts. Specifically, our key contributions are threefold:

\begin{enumerate}
	\item We propose a score-based conditional objective and architecture designed to effectively learn spatial attributes of interacting 3D objects in a constrained indoor environment. In contrast to previous approaches~\cite{diffuscene,atiss}, we disentangle the design space and reduce the model's prediction to a minimal representation consisting solely of oriented 3D bounding boxes, taking as conditioning input the room's floor plan and list of object semantic categories.
	
	\item We propose a set of approaches which allows a model trained following our method to be flexibly employed at test time to perform several user-driven tasks enabling object or attribute-level control. In particular, we demonstrate strong capabilities on controllable scenarios such as scene re-arrangement or room completion, from a single trained network.
	
  \item Finally, we introduce a novel \textit{Self Score Evaluation} (SSE) procedure, which enables 3D scene synthesis by selecting the set of inputs provided by external sources, such as a LLM, that lead to the most realistic layouts.
	
\end{enumerate}

We exhibit our model's capabilities across a wide range of experimental scenarios and report state-of-the art 3D layout generation and scene synthesis performance.

\begin{figure}[t]
  \centering
  \includegraphics[width=1.\textwidth]{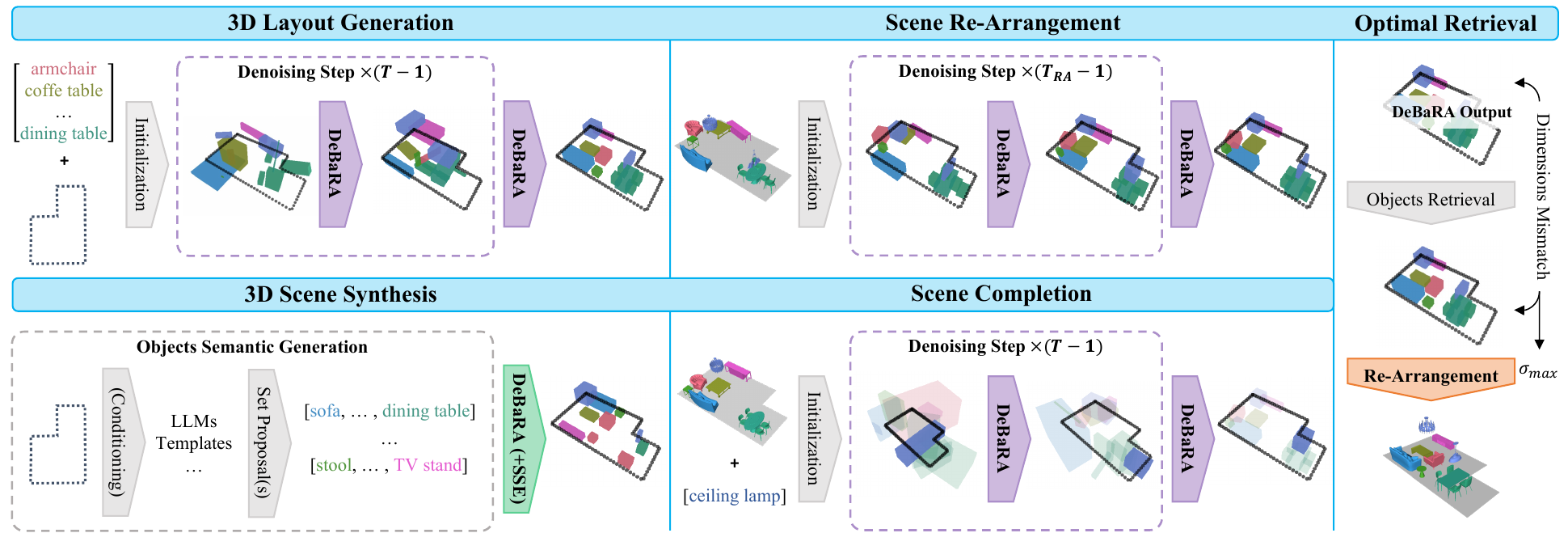}
  \caption{\textbf{Application scenarios overview.} Besides generating diverse and realistic 3D indoor layouts, a single trained DeBaRA model can be employed to execute several related tasks by tweaking the initial sampling noise level \(\sigmamax\) and/or performing object or attribute-level layout inpainting. Our novel SSE procedure enables 3D Scene Synthesis capabilities by efficiently selecting conditioning semantics from external sources using density estimates provided by the pretrained model.}
  \label{fig:scenarios}
\end{figure}

\section{Related Work}

\paragraph{Score-based Generative Models}

By smoothly perturbing training examples with noise, Diffusion Models map a complex data distribution to a known Gaussian prior from which they sample back via iterative denoising using a neural network trained over multiple noise levels. This family of generative models has been motivated by several theoretical foundations over the past years: DDPMs~\cite{ddpm, iddpm} parameterize the diffusion process as a discrete-time Markov chain, as opposed to \textit{continuous-time} approaches~\cite{song2020score,song2021maximum}. The seminal EDM~\cite{karras,karras2024analyzing} training and sampling settings later unified previous methods into an improved ideal framework defined by a set of interpretable parameters. Originally motivated by image generation, diffusion models have demonstrated impressive capabilities on various conditional tasks such as text-to-image synthesis~\cite{glide,ldm}, image-to-image generation from various 2D input modalities~\cite{ldm,controlnet,3dim}, text-to-3D asset creation~\cite{dreamfusion,instant3d} or environment-aware human motion synthesis~\cite{huang2023diffusion,chois}. Relevant to our work, diffusion models have been applied to the generation of point clouds~\cite{luo2021diffusion,vahdat2022lion} and other geometric representations involving 3D coordinates~\cite{xu2024brepgen}.

\paragraph{Lifting Pretrained Diffusion Models}

Knowledge of trained diffusion models can be leveraged in various settings including content inpainting\cite{repaint,huang2023diffusion}, score distillation~\cite{dreamfusion}, exact likelihood computation~\cite{song2020score,karras} or teacher-student distillation~\cite{progressive,meng2023distillation}. More relevant to our work, image-domain diffusion priors have demonstrated compelling performance in discriminative tasks including zero-shot image classification~\cite{li2023your,clark2024text,chen2024your} and segmentation~\cite{peekaboo}. More precisely, Diffusion Classifiers assign a label, from a finite set of possible classes \(\left\{\cc_i\right\}_{i=1}^N\) to an observed sample \(\xx_0\) by computing class-conditional density estimates from a pretrained diffusion model under the assumption of a uniform prior \(p\left(\cc_i\right)=1/N\). In practice, this is done by, for each class, iteratively adding noise to the observed sample \(\xx_0\) and computing a Monte Carlo estimate of the expected reconstruction loss using the class-conditioned model.

\paragraph{Controllable 3D Scene Synthesis}

Synthesizing indoor 3D layouts from a partial set of information or constraints has come in various settings depending on provided vs.~predicted entities and enabled control granularity. A prolific line of research has been adopting intermediate 3D scene representations such as graphs~\cite{li2019grains,fastsynth,planit,wonkagraph,commonscenes,gao2023scenehgn,lin2023instructscene}, furniture matrices~\cite{matrix} or multi-view images~\cite{3Dsp2Ds}. Autoregressive furnishing approaches~\cite{sceneformer,atiss} have been supplemented by object attribute-level conditioning~\cite{cofs,cliplayout} and additional ergonomic constraints~\cite{layoutenhancer}. However, their \textit{one object at a time} strategy does not comprehensively capture complex relationships between all the interacting elements and is known to easily fall into local minima in which new items fail to be accurately inserted to the current configuration. Lately, methods have unfolded LLMs double-edged capabilities in this area~\cite{layoutgpt,holodeck} as they excel at generating sensible furniture descriptions while struggling in accurately arranging them in the 3D space, which~\cite{anyhome} addresses by introducing a costly refinement stage. In the light of that, LLMs appear to be ideal candidates to supplement a specialized 3D layout generation model.

\paragraph{Denoising Indoor Scenes}

Previous methods have explored diffusion-based approaches in the context of 3D scene synthesis. Pioneering their usage, LEGO-Net~\cite{legonet} performs scene re-arrangement (i.e., recovering a \textit{clean} object layout from a \textit{noisy} one) in the 2D space using a transformer backbone that is not noise-conditioned, which we argue is the root cause of its main limitations. PhyScene\cite{yang2024physcene} augment diffusion-based 3D scene synthesis with additional physic-based guidance to enable practical embodied agent applications. Most relevant to our work, DiffuScene~\cite{diffuscene} achieves 3D scene synthesis by fitting a DDPM~\cite{ddpm} on stacked 3D object features, resulting in a high-dimensional composite distribution that is hard to learn and interpret. It does not enforce spatial configurations over other predicted features. More importantly, its generative process is not conditioned on the room's floor plan (i.e., bounds) that constrains objects to be placed within a restricted area.

\section{Method}
\label{section:method}

\subsection{3D Scene Representation}

Our method is based on encoding the state of a 3D indoor scene \(\mathcal{S}\) that is defined by a floor plan (i.e., bounds) \(\mathcal{F}\) and an unordered set of \(N\) objects \(\mathcal{O} = \{o_1, \dots, o_N\}\), each being modeled by its typed 3D bounding box \(o_i=\{\xx_i, \cc_i\}\) where \(\cc_i \in {\{0,1\}}^k\) is the one-hot encoding of the semantic category among \(k\) classes and \(\xx_i = (\pp_i, \rr_i, \dd_i) \in  \mathbb{R}^8\) comprises 3D spatial attributes. More specifically, \(\pp_i \in \mathbb{R}^3\) denotes the object's center coordinate position, \(\rr_i = (\cos\theta_i, \sin\theta_i)\in \mathbb{R}^2\) is a continuous encoding of the rotation of angle \(\theta_i\) around the scene's vertical axis~\cite{continuity} and \(\dd_i \in \mathbb{R}^3\) is the dimension.

\subsection{Diffusion Framework and Architecture}
\label{section:method-framework}

\begin{figure}[t!]
  \centering
  \includegraphics[width=1.\textwidth]{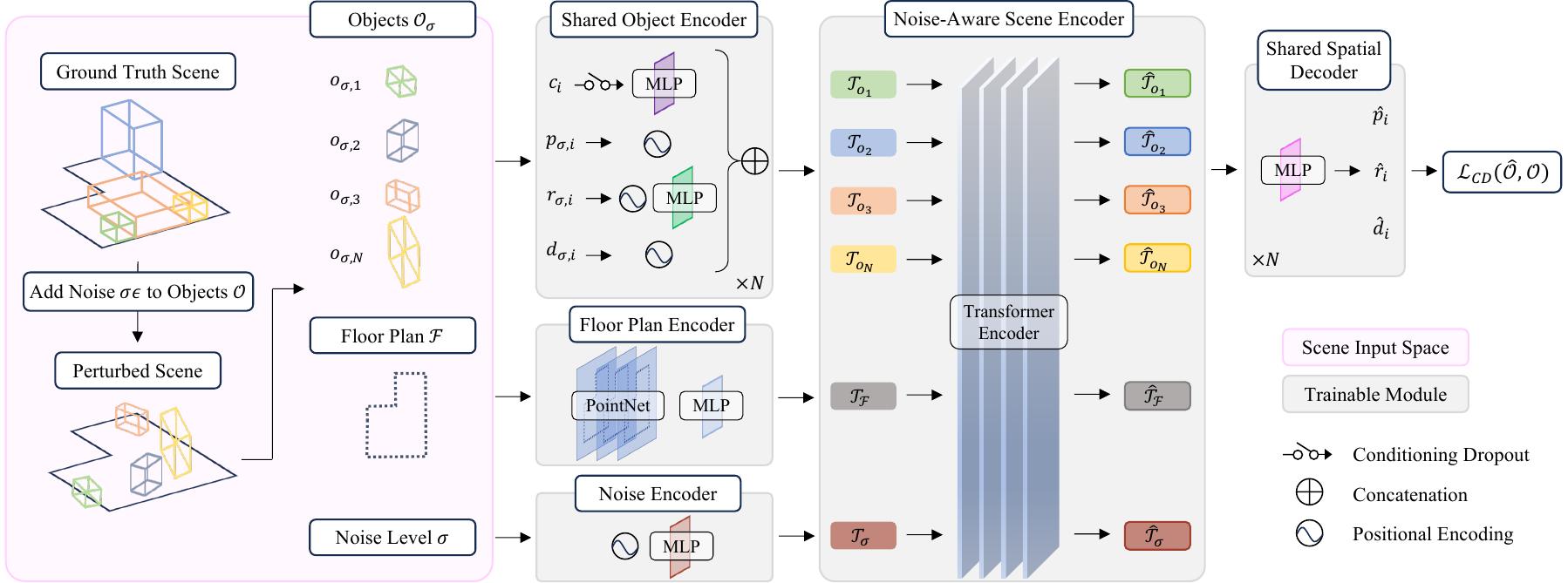}
  \caption{\textbf{DeBaRA architecture and training overview.} At each iteration, 3D bounding boxes parameters \((\pp, \rr, \dd)\) of indoor scene's objects \(\mathcal{O}\) are perturbed with Gaussian noise \(\sigma \boldsymbol{\epsilon}\). The floor plan \(\mathcal{F}\), noise level \(\sigma\) and resulting objects \(\mathcal{O_\sigma}\) are processed by respective encoders to form an unordered set of representations \(\mathcal{T}\) fed as input to a transformer encoder. Novel object embeddings \(\hat{\mathcal{T}}_{o}\) are finally decoded back to their predicted \textit{clean} spatial configuration \((\hat{\pp}, \hat{\rr}, \hat{\dd})\). Trainable modules are optimized by minimizing a semantic-aware Chamfer loss. Input object categories \(\cc\) are randomly dropped to model both the class-conditional and unconditional 3D layout distributions.}
  \label{fig:archi}
\end{figure}

We describe in this section our score-based layout generation framework, relevant design choices and network architecture, that are summarized in Figure~\ref{fig:archi}. Remarkably, unlike previous approaches~\cite{diffuscene,atiss,cofs} that output a range of attributes lying in different spaces, we focus on accurately modeling 3D spatial layouts of bounded indoor scenes from a set of input object categories. 

\paragraph{Learning 3D spatial configurations from object semantics}

We adopt a diffusion-based approach to yield a conditional generation model that outputs 3D object spatial features \({\{\xx_i\}}_{i=1}^N\) from an input floor plan and set of semantic categories \(\yy = (\mathcal{F},\cc)\) with \(\cc = \{\cc_i\}_{i=1}^{N}\). During training, 3D spatial attributes are perturbed with Gaussian noise \(\boldsymbol{\epsilon}\sim \mathcal{N}(\boldsymbol{0}, \boldsymbol{\mathrm{I}})\) at various noise levels (i.e., magnitudes) \(\sigma\). A trainable noise-conditioned denoiser model \(D_\theta(\xx_\sigma; \yy, \sigma)\) maps \textit{noisy} spatial attributes \(\xx_\sigma = \xx + \sigma\boldsymbol{\epsilon}\) to their \textit{clean} counterparts \(\hat{\xx}\approx \xx \in \mathbb{R}^{N \times 8}\). 

We notice that each object spatial attribute has an individual real-world interpretation (e.g, \(\pp\) and \(\dd\) can be expressed in meters, \(\rr\) in degrees). To preserve their measurable nature at intermediate perturbed configurations \(\xx_\sigma\), we want our diffusion parameterization to support a continuous range of noise levels, correlated to the scale of the input signal. This will be particularly convenient at test time (see Section \ref{section:method-applications}). To guarantee both properties, we adapt the score-based EDM~\cite{karras} framework. In our context, this formulation is more natural than the DDPM framework employed by previous work~\cite{diffuscene}. The latter is based on \textit{discretizing} noise levels and does not offer a straightforward interpretability of the scene's state at arbitrary timesteps. Our parameterization is further detailed in Appendix~\ref{sec:sup:framework}.

\paragraph{Estimating the unconditional layout density}

Inspired by \emph{classifier-free guidance}~\cite{cfg} in the image domain, we model both the class-conditional density \(p_\theta\bigl(\xx | \mathcal{F}, \cc\bigl)\) and the unconditional density  \(p_\theta\bigl(\xx | \mathcal{F}, \emptyset)\) by a single network of parameters \(\theta\). At each training iteration, we perform \emph{conditioning dropout} on the set of semantic categories, by setting \(\cc = \emptyset\) with probability \(\pdrop\) else \(\{\cc_i\}_{i=1}^{N}\). We found that this mechanism helps to reduce overfitting of the training layouts \(p_{\stext{data}}(\xx)\) and enables novel capabilities that we introduce in Section \ref{section:sse}.

\paragraph{Denoiser Network Architecture}

Our lightweight architecture is inspired by previous work~\cite{legonet} to which we make key changes. Similar to~\cite{atiss} and~\cite{diffuscene}, we use a shared object encoder in order to obtain per-object token \(\mathcal{T}_{o_i}\) as a concatenation of the object \(o_i\) attributes embedded by sinusoidal positional encoding and linear layers. Following \cite{legonet}, we uniformly sample \(P\) points on the edges of the floor plan and feed them into a PointNet~\cite{qi2017pointnet} model, resulting in a floor token \(\mathcal{T}_{\mathcal{F}}\). This choice of feature extractor backbone is natural as it allows to maintain all the input scene's spatial features in a common 3D space. Importantly, a noise token \(\mathcal{T}_\sigma\) is computed from the current noise level \(\sigma\), making our architecture \textit{noise-aware}, i.e., able to denoise layouts \(\xx_\sigma\) at any perturbation magnitude.
 
All the previously encoded tokens form a sequence \(\mathcal{T} = \{\mathcal{T}_{\mathcal{F}}, \mathcal{T}_{\sigma}, \mathcal{T}_{o_i}, \dots, \mathcal{T}_{o_N}\}\) from which a global scene encoder \(T_\theta\) computes rich representations \(\hat{\mathcal{T}}\). We design the method without any token ordering and use padding mask for scene with fewer objects than the transformer capabilities. A final shared decoder MLP takes as input object tokens \(\{\hat{\mathcal{T}}_{o_i}\}_{i=1}^{N}\) and returns denoised spatial attribute values \(\hat{\xx} = \{(\hat{\pp}_i, \hat{\rr}_i, \hat{\dd}_i)\}_{i=1}^{N}\). We provide implementation details on the denoiser in Appendix~\ref{sec:sup:imp:network}.

\subsection{3D Spatial Objective}
\label{section:method:objective}

Our noise-conditioned model \(D_\theta\) is optimized towards a novel semantic-aware Chamfer Distance objective that does not penalize permutation of 3D bounding boxes sharing the same semantic category between the predicted scene objects layout \(\hat{\mathcal{O}}\) and the ground truth one \(\mathcal{O}\):
\begin{equation}
  \mathcal{L}_{CD}(\hat{\mathcal{O}},\mathcal{O}) = \frac{1}{2N}\left(\sum_{\hat{o}\in \hat{\mathcal{O}}}\min_{o\in \mathcal{O}}l(\hat{o},o) + \sum_{o\in \mathcal{O}}\min_{\hat{o}\in \hat{\mathcal{O}}}l(\hat{o},o)\right),
\end{equation}
\begin{equation} 
  \quad \text{where} \quad l(\hat{o}, o) = \|\hat{\xx}-\xx\|_2^2 + \kappa \bigl(1-\delta_{\cc}(\hat{o},o)\bigr).
\end{equation}
Here, \(\kappa\) is a large value so that a significant penalty is applied to objects that do not share the same semantic category \(\cc\), preventing them to be returned by the \(\min\) operator.

We can finally rewrite the usual score-based training objective~\cite{song2020score,karras} as:
\begin{equation}
  \mathbb{E}_{p_{\stext{data}}(\xx),\boldsymbol{\epsilon}, \sigma}\bigl[\lambda(\sigma)\mathcal{L}_{CD}(D_\theta(\xx + \sigma \boldsymbol{\epsilon}; \yy, \sigma), \xx)\bigl]
\label{eq:diffusion-loss}
\end{equation}
where \(\lambda(\sigma)\) is a noise-dependent loss weighting function.

\subsection{Self Score Evaluation}
\label{section:sse}

\begin{algorithm}[b!]
  \caption{Self Score Evaluation}
  \begin{algorithmic}[1]
  \Require
  a diffusion prior \(D_\theta\) trained with conditioning dropout and by optimizing \(\mathcal{L}_{CD}\)
  \Ensure
  conditioning candidates \(\{\cc_j\}_{j=1}^C\), number of score evaluation trials \(\Tsse\)
  \State \textbf{sample} \(\xx_{j}\sim p_\theta(\xx | \mathcal{F}, \cc_j)\) for each candidate \(\cc_j\) using iterative sampling
  \State \textbf{initialize} \(\texttt{scores}[\cc_j]=\texttt{list}()\) for each \(\cc_j\)
  \For{trial \(t = 1, \ldots, \Tsse\)}
  \State \textbf{sample} \(\sigma \sim \mathcal{N}(0, \sigma_s); \boldsymbol{\epsilon} \sim \mathcal{N}(\boldsymbol{0},\boldsymbol{\mathrm{I}})\)
  \For{candidate \(\cc_k\), sample \(\xx_k\)}
  \State \(\texttt{scores}[\cc_k].\texttt{append}(\mathcal{L}_{CD}[D_{\theta}(\xx_k+\sigma\boldsymbol{\epsilon},;\,\mathcal{F},\emptyset,\sigma), \xx_k])\)
  \EndFor
  \EndFor\\
  \Return \(\arg\,\min_{\cc_j}\) \texttt{mean(scores}[\(\cc_j\)])
  \end{algorithmic}
  \label{alg:sse}
\end{algorithm}

While specifying complete conditioning information such as the set of object semantics \(\cc\) could be tedious, it can be \textit{provided} by either a LLM or a separately trained sequence generation model. However, using independent models is inherently suboptimal since it does not guarantee that the generated conditioning input will be aligned with the score model knowledge. As a result, we propose a novel method to select conditioning inputs that are attuned with the model's capabilities.

More specifically, we evaluate a finite set of \(C\) object semantic categories \emph{candidates}, where each candidate is associated to a 3D spatial layout sampled from the learned conditional density, i.e.,
\begin{equation}
  \text{candidates} = \biggl\{\bigl({\cc_{j},\xx_{j}\sim p_\theta\bigl(\xx | \mathcal{F}, \cc_j})\bigl)\biggl\}_{j=1}^C
\label{eq:sse-candidates}
\end{equation}
Then, the optimal conditioning candidate \(\cc^{*}\) is derived from a density estimate of its corresponding 3D spatial layout \(\xx^{*}\) provided by the unconditional network:
\begin{equation}
  \xx^{*} = \arg\,\min_{\xx_i}\,\mathbb{E}_{\boldsymbol{\epsilon}, \sigma}\bigl[\mathcal{L}_{CD}\{D_{\theta}(\xx_i+\sigma\boldsymbol{\epsilon}\,;\,\mathcal{F},\emptyset,\sigma), \xx_i\}\bigl]
  \label{eq:sse-mce}
\end{equation}
In practice, we compute an unbiased Monte Carlo estimate of each candidate expectation using \(\Tsse\) fixed \((\sigma, \boldsymbol{\epsilon})\) pairs. Although similar in some aspects, SSE fundamentally differs from diffusion classifiers~\cite{li2023your} as in our case, the uniform assumption over conditioning probabilities does not hold. Indeed, in our setting some input signals cannot lead to a plausible arrangement at all. As a result, density estimates of observed samples generated by the class-conditioned model are computed using the unconditional one, while diffusion classifiers compute density estimates of a single observed sample using the class-conditioned model. The SSE procedure is detailed in Algorithm~\ref{alg:sse}. It is further illustrated and discussed in Appendix~\ref{sec:sup:sse}.

\begin{figure}[!t]
  \centering
  \begin{subfigure}[b]{0.3\textwidth}
    \centering
    \includegraphics[width=0.55\textwidth]{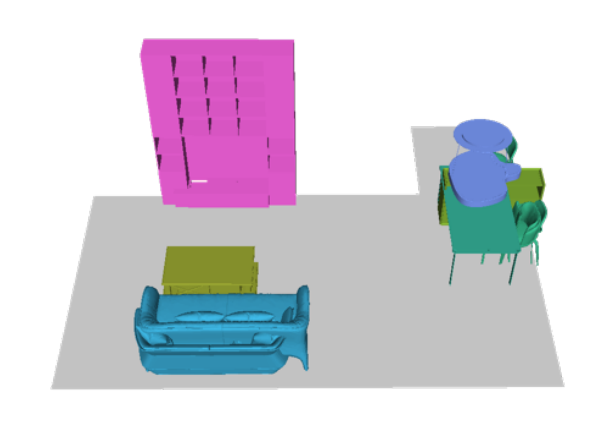}\vspace{4pt}
    \includegraphics[width=0.65\textwidth]{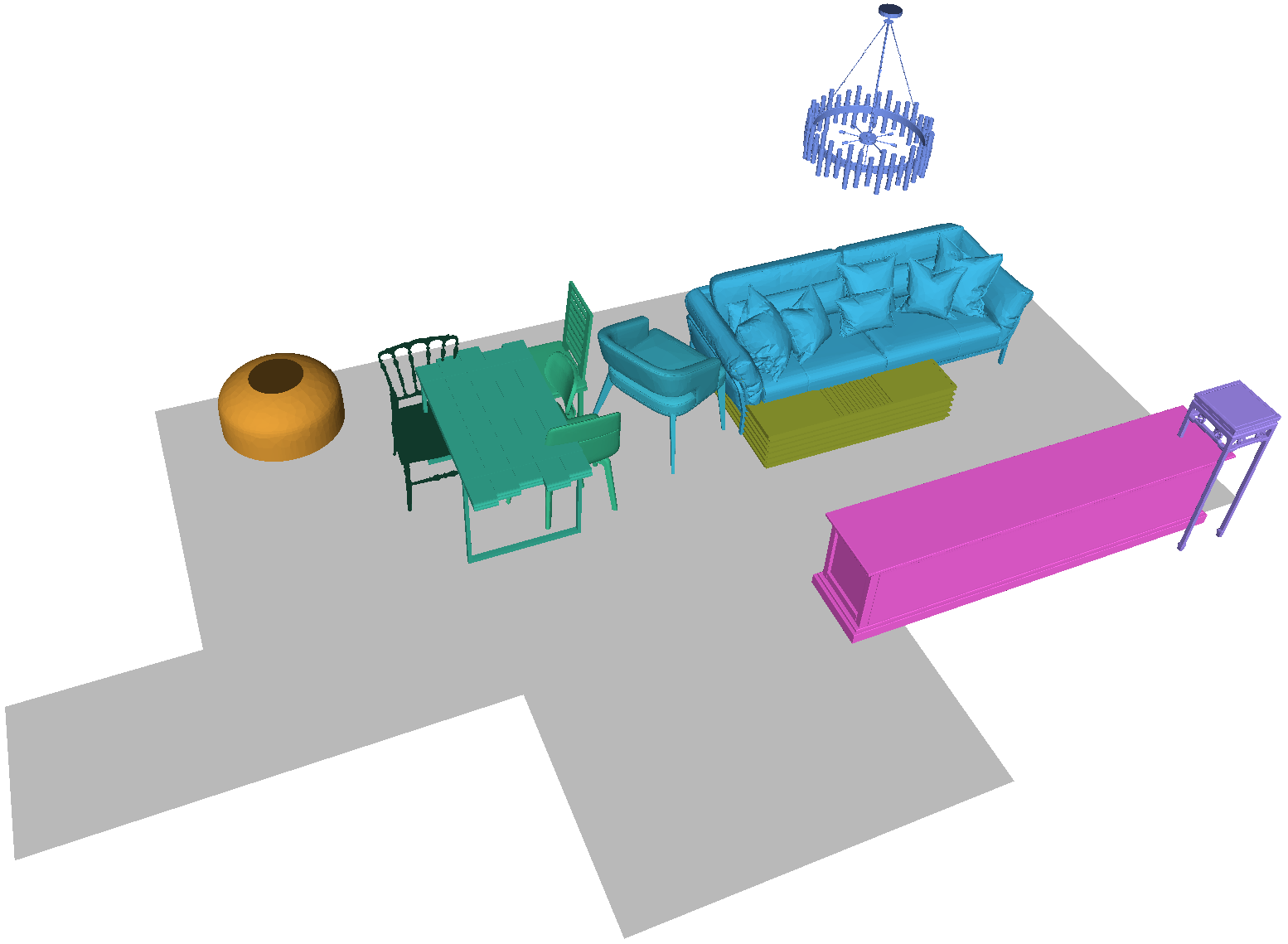}\vspace{2pt}
    \includegraphics[width=0.77\textwidth]{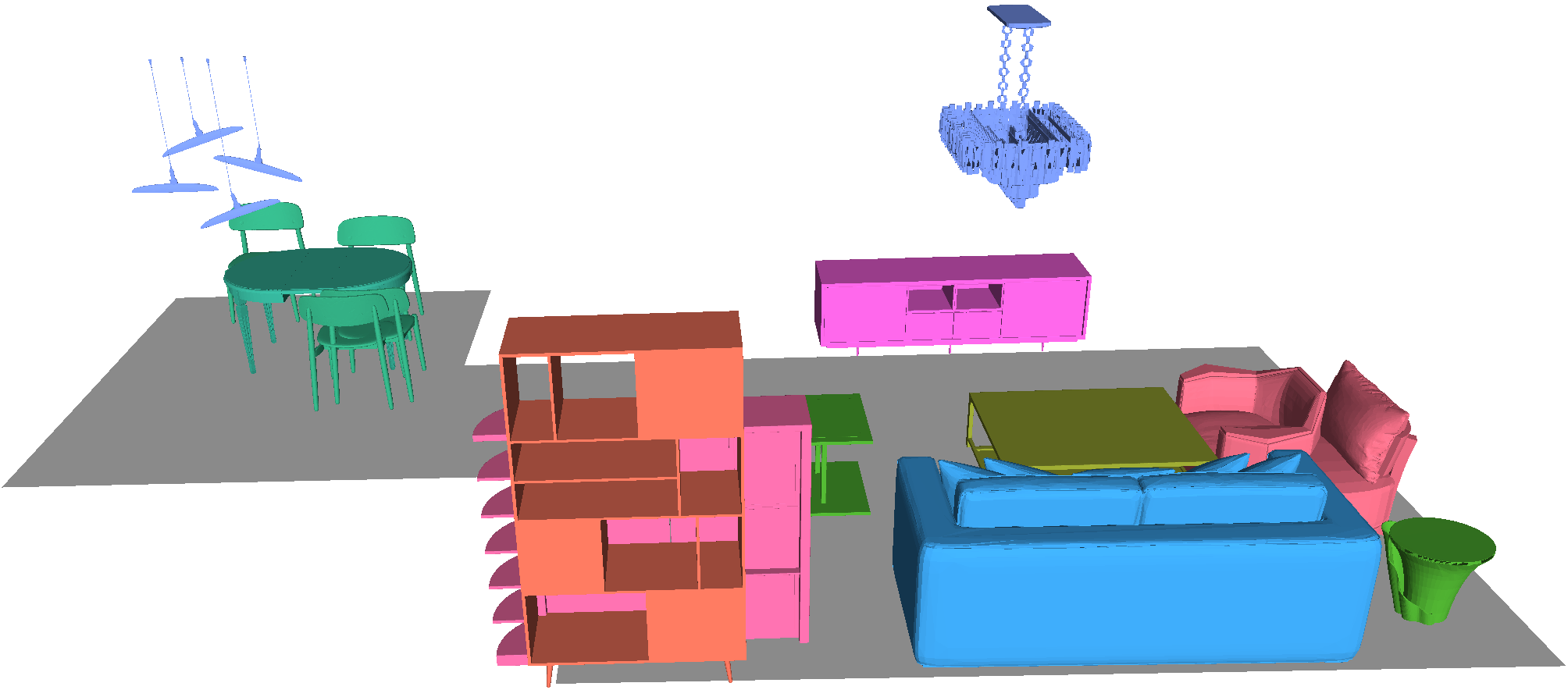}
    \vspace{4pt}
    \caption{ATISS~\cite{atiss}}
  \end{subfigure}
  \hfill
  \vrule
  \hfill
  \begin{subfigure}[b]{0.3\textwidth}
    \centering
    \includegraphics[width=0.55\textwidth]{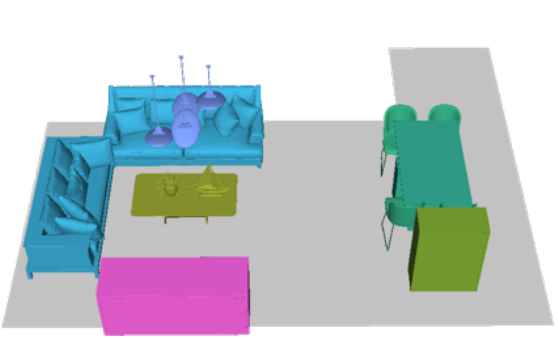}\vspace{4pt}
    \includegraphics[width=0.65\textwidth]{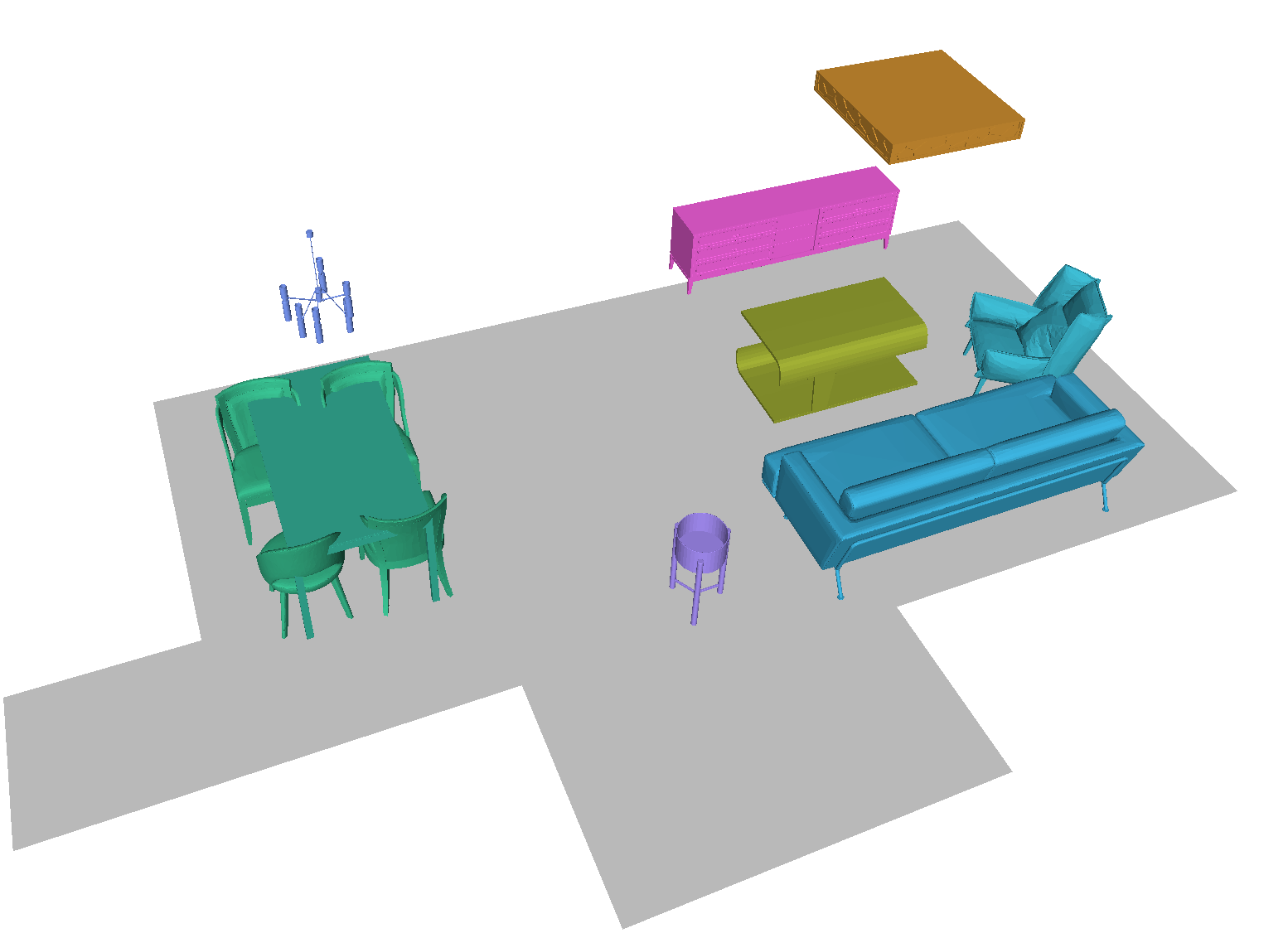}\vspace{2pt}
    \includegraphics[width=0.77\textwidth]{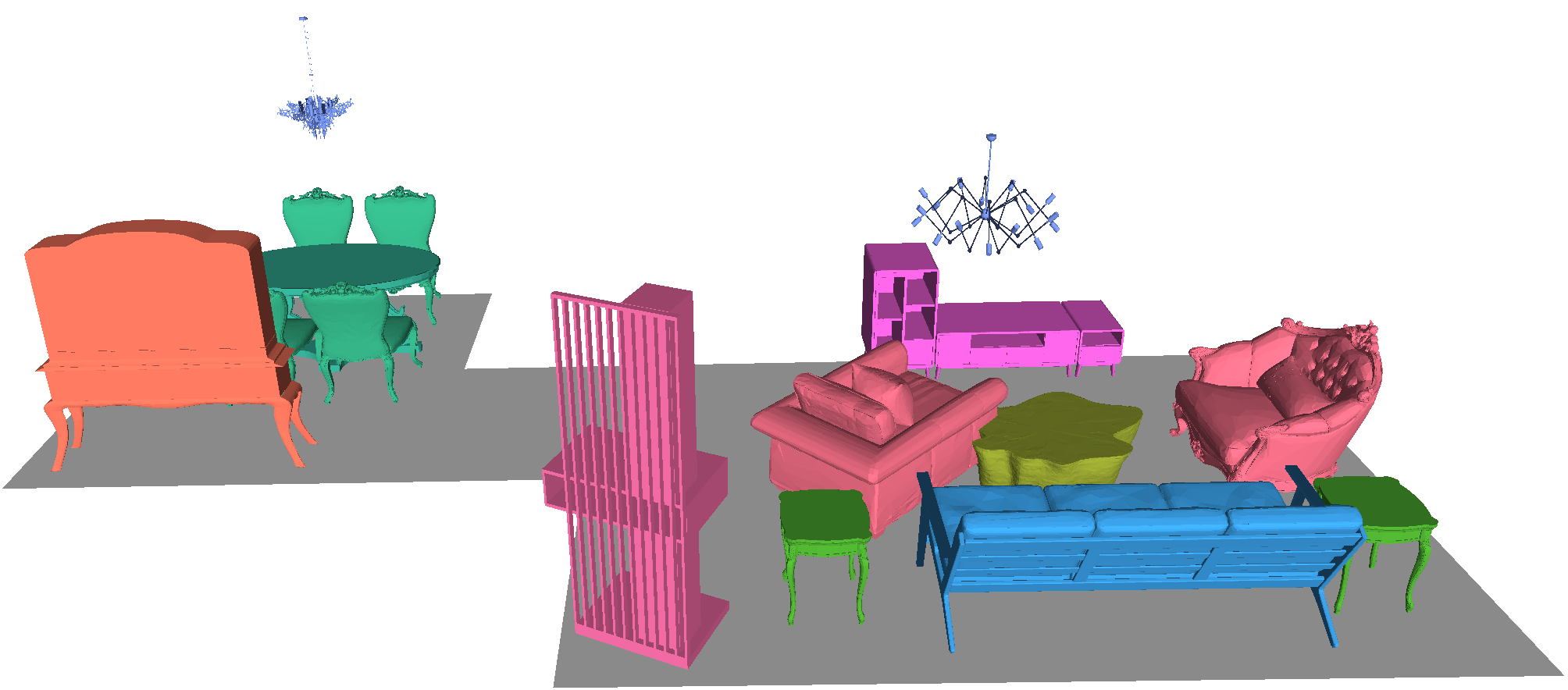}
    \vspace{4pt}
    \caption{DiffuScene~\cite{diffuscene}}
  \end{subfigure}
  \hfill
  \vrule
  \hfill
  \begin{subfigure}[b]{0.3\textwidth}
    \centering
    \includegraphics[width=0.55\textwidth]{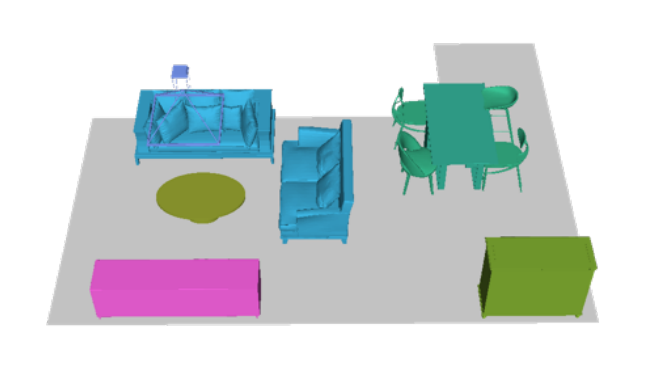}\vspace{4pt}
    \includegraphics[width=0.65\textwidth]{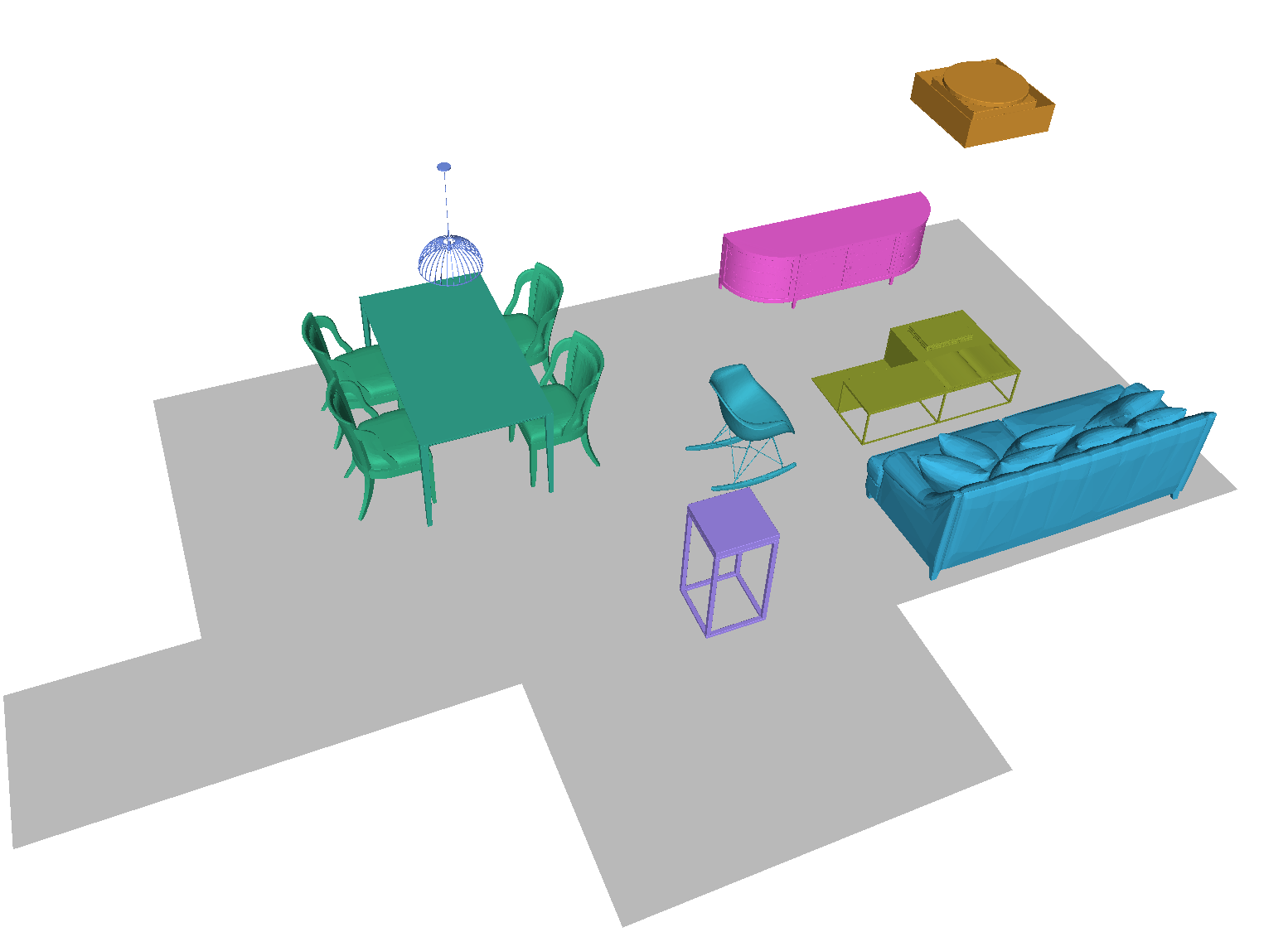}\vspace{2pt}
    \includegraphics[width=0.77\textwidth]{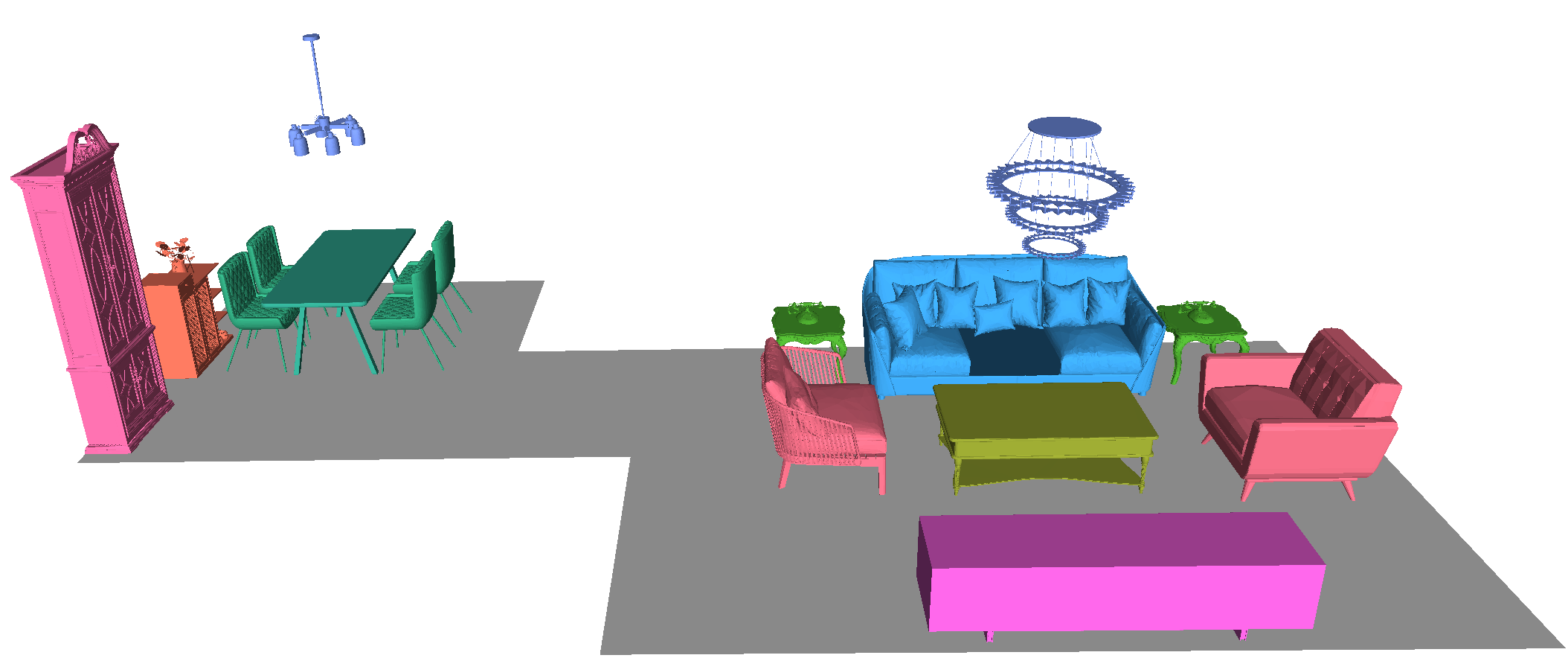}
    \vspace{4pt}
    \caption{\textbf{DeBaRA}}
  \end{subfigure}
  \caption{We compare our method with established baselines for generating a 3D layout from a floor plan and set of object categories. DeBaRA produces less failure cases while consistently generating regular arrangements within the room's bounds.}
  \label{fig:arrangegen}
\end{figure}

\subsection{Application Scenarios}
\label{section:method-applications}

As shown in Figure~\ref{fig:scenarios}, a single trained DeBaRA model can be used at test time to perform multiple downstream interactive applications. Usual generation procedures, such as EDM 2\textsuperscript{nd} order stochastic sampler~\cite{karras} can be applied using our trained denoiser to generate novel 3D layouts via \(T\text{-step}\) iterative denoising at discretized noise levels \(\sigma_0 = \sigmamax > \ldots > \sigma_T = 0 \).

In particular, several applications can be performed by inpainting~\cite{repaint}, i.e., predicting missing spatial features from those specified (i.e., fixed) in the input layout \(\xx \in \mathbb{R}^{N \times 8} \). To do so, we introduce a binary mask \(\mathbf{m} \in \{0,1\}^{N \times 8}\) specifying values to retain from the input. The predicted layout at any sampling iteration \(t\) can be expressed as:
\begin{equation}
  \tilde{\xx}_{\sigma_t} = \hat{\xx}_{\sigma_t} \odot (1 - \mathbf{m}) + \xx_{\sigma_t} \odot \mathbf{m}
\end{equation}
\paragraph{3D Layout Generation}

Novel and diverse 3D layouts can be generated from an input set of semantic categories \(\cc\) and a floor plan \(\mathcal{F}\) by sampling from a high initial noise level \(\sigmamax >> \sigmadata\), arbitrarily initialized 3D spatial features \(\xx_{\sigma_0}\) and with \(\mathbf{m} = \mathbf{0}_{N \times 8}\).

\paragraph{3D Scene Synthesis}

DeBaRA can perform 3D scene synthesis via 3D layout generation from semantic categories provided by external sources such as a LLM~\cite{layoutgpt}. Input conditioning candidates can further be optimally selected using the Self Score Evaluation procedure.

\paragraph{Scene Completion}

Additional objects \(o_a\) can be inserted to an existing scene partially furnished with \(k\) objects \(o_e\). To do so, their 3D spatial attributes \(\xx_a\) are inpainted from the existing ones \(\xx_e\) with \(D_\theta\) conditioned on the updated set of semantic categories \(\cc = \cc_e \| \cc_a\) using \(\mathbf{m}(i,j) = \mathbf{1}_{\{i \le k\}}\).

\paragraph{Re-arrangement}

In the context of scene synthesis, re-arrangement~\cite{legonet} consists in recovering the closest \textit{clean} spatial configuration of existing objects from a \textit{messy} one, which has practical applications in robotics~\cite{batra2020rearrangement}. DeBaRA can perform re-arrangement by sampling from an initial noise \(\sigmamax\) that depends on the scene perturbation magnitude. During denoising, object positions and rotations \((\pp, \rr) \in \mathbb{R}^{N \times 5}\) are inpainted from the known object dimensions using \(\mathbf{m}(i,j) = \mathbf{1}_{\{j > 5\}}\).

\paragraph{Optimal Object Retrieval}

3D scene synthesis systems depend on external 3D asset databases for furnishing rooms. For each object of semantic class \(\cc\), a textured furniture is retrieved by minimizing the mismatch with the generated dimension \(\dd_{\sigma_T}\). This is inherently suboptimal as the resulting scene quality is limited by the size of the external database. To overcome this issue, we introduce a post-retrieval refinement stage by performing additional \textit{re-arrangement} steps starting from a noise level \(\sigmamax\) derived from the mismatch between generated and retrieved object dimensions.

\paragraph{Generation from Coarse Specifications}

We propose a time-dependent masking approach to synthesize layouts from \textit{rough} input spatial features (i.e., instead of \textit{exact} ones), that are adjusted in the late denoising iterations. To indicate approximate e.g., object dimensions, we set, at any sampling step \(t\), \(\mathbf{m}(i,j) = \mathbf{1}_{\{j > 5 \text{ and } t < T_s\}}\). The denoising step \(T_s\) from which \(\mathbf{m}\) is \textit{relaxed} (i.e., set to \(\mathbf{0}\)) can be derived from its corresponding noise level \(\sigma_{T_s}\) and the precision of specified input features.

\section{Experiments}
\label{sec:experiments}

\begin{figure}[t]
  \centering
  \begin{subfigure}[b]{0.18\textwidth}
    \centering
    \includegraphics[width=0.92\textwidth]{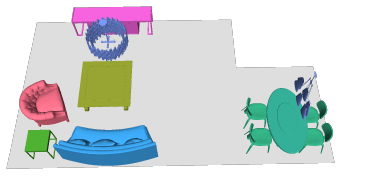}
    \includegraphics[width=0.92\textwidth]{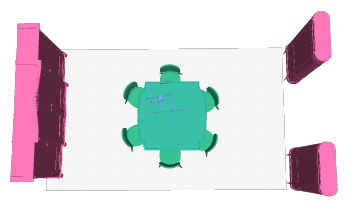}
    \caption{Ground Truth}
  \end{subfigure}
  \hfill
  \begin{subfigure}[b]{0.18\textwidth}
    \centering
    \includegraphics[width=0.92\textwidth]{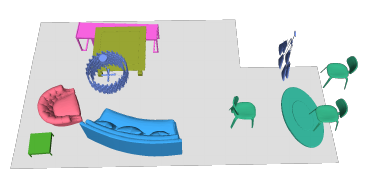}
    \includegraphics[width=0.92\textwidth]{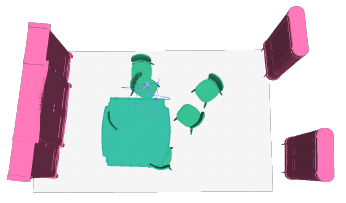}
    \caption{Noisy}
  \end{subfigure}
  \hfill
  \begin{subfigure}[b]{0.18\textwidth}
    \centering
    \includegraphics[width=0.92\textwidth]{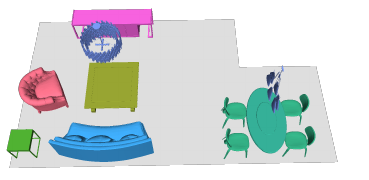}
    \includegraphics[width=0.92\textwidth]{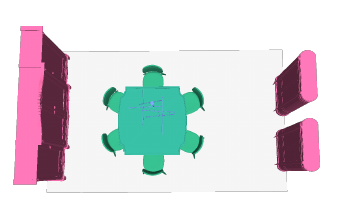}
    \caption{\textbf{Re-arranged}}
  \end{subfigure}
  \hfill
  \vrule
  \hfill
  \begin{subfigure}[b]{0.18\textwidth}
    \centering
    \includegraphics[width=0.92\textwidth]{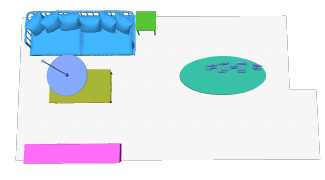}
    \includegraphics[width=0.92\textwidth]{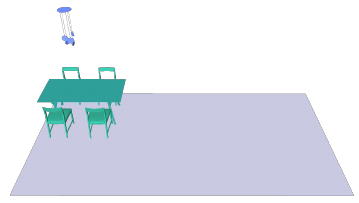}
    \caption{Partial}
  \end{subfigure}
  \hfill
  \begin{subfigure}[b]{0.18\textwidth}
    \centering
    \includegraphics[width=0.92\textwidth]{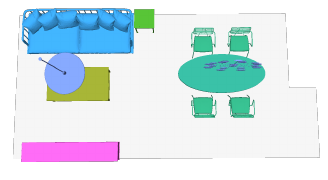}
    \includegraphics[width=0.92\textwidth]{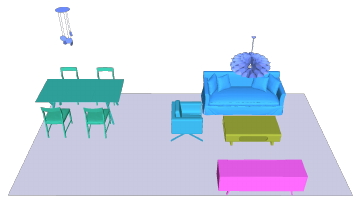}
    \caption{\textbf{Completed}}
  \end{subfigure}
  \caption{Qualitative results on \textbf{scene re-arrangement} (left) and \textbf{completion} (right). DeBaRA is able to recover a plausible layout from a messy one, and to finely take into account initial configurations.{\protect}}  
  \label{fig:arrange-complete}
\end{figure}

In this section, we provide a comprehensive experimental evaluation of DeBaRA that we compare with established baselines from different model families. We also demonstrate the capabilities of our approach in various practical scenarios, enabling a wide range of applications.

\paragraph{Datasets}

Our experiments are conducted on the 3D-FRONT~\cite{3dfront} synthetic indoor layouts, furnished with assets from 3D-FUTURE~\cite{3dfuture} that we use as the object retrieval database. Out of the available room types in the dataset, we independently consider living rooms and dining rooms which are more densely furnished and feature complex floor plans. We follow the preprocessing from ATISS~\cite{atiss}, leading respectively to \(2338\)/\(587\) and \(2071\)/\(516\) train/test splits.

\paragraph{Baselines}

We compare DeBaRA with ATISS~\cite{atiss} autoregressive transformer and DiffuScene~\cite{diffuscene} denoising network. To ensure a fair comparison with our method, we retrained both models with floor plan conditioning on each 3D-FRONT subset using their official implementations. To perform 3D arrangement generation with DiffuScene, we implemented DDPM inpainting~\cite{repaint} of object spatial features from their known semantic categories. Additionally, we report experimental results obtained by LayoutGPT~\cite{layoutgpt} that we implemented with a Llama-3-8B backbone~\cite{llama3} that we also use to provide semantic categories in scene synthesis scenarios. Following the paper, we perform prompting with \textit{supporting examples}: for each test scene, we retrieve top-\(k\) samples from the training set that have the most similar floor plan and include their spatial configuration as few-shot exemplars. Note that LayoutGPT adopts a training-free approach and is therefore not directly comparable to our method. However, we show how it can be used \textit{alongside} a specialized model such as DeBaRA. Full implementation details and LLM prompting strategies are reported in Appendix~\ref{sec:sup:baselines}.

\paragraph{Evaluation Metrics}

We follow previous work~\cite{fastsynth,atiss,diffuscene,layoutgpt,sceneformer} and evaluate the realism and diversity of generated arrangements by reporting the \(256^2\) Fréchet Inception Distance (FID)~\cite{heusel2017gans}, Kernel Inception Distance (KID \(\times 1,000\))~\cite{binkowski2018demystifying} and Scene Classification Accuracy (SCA, values closer to \(50\%\) are better) computed on top-down orthographic renderings. Resulting projections feature the scene's floor plan and objects colored according to their semantic class~\cite{diffuscene}. The generation spatial validity is further assessed by reporting the cumulated out of bounds objects area (OBA, in \(m^{2}\)). Related indicators are provided and discussed in Appendix~\ref{sec:sup:res:bound}. Metrics are computed across each test subset, for which we generate the same number of scenes as the number of \textit{real} ones.

\subsection{3D Layout Generation}

The primary task of DeBaRA is to generate diverse and valid 3D layouts within a given floor plan and a list of object semantics. We showcase qualitative generation results and comparisons in Figure~\ref{fig:arrangegen}. As highlighted by previous work~\cite{legonet,diffuscene}, denoising-based methods better capture the interplay between interacting objects. We also observe that DeBaRA largely outperforms baselines at respecting the scene's bounds while consistently producing more natural arrangements. These observations are quantitatively verified in Table~\ref{tab:arrangegen} and visualized in Figure~\ref{fig:arrangegen}.

\begin{table*}[!h]
  \centering
  \caption{Quantitative experiment results on \textbf{bounded 3D layout generation} (providing a floor plan and a list of object semantic categories). We compare our method against other learning-based approaches and additionally indicate results obtained from a training-free LayoutGPT.
  }
  \resizebox{\textwidth}{!}{%
  \begin{tabular}{l l cccc c cccc}
  \toprule
      & \multicolumn{1}{c}{\multirow{2.5}{*}{\textbf{Methods}}} & \multicolumn{4}{c}{\textbf{Living Rooms}} && \multicolumn{4}{c}{\textbf{Dining Rooms}}\\

      \cmidrule(lr){3-6} \cmidrule(lr){8-11}

      & & \multicolumn{1}{c}{FID ($\downarrow$)} & \makecell{KID ($\downarrow$)} & {SCA (\%)} & OBA ($\downarrow$) && \makecell{FID ($\downarrow$)} & KID ($\downarrow$) & {SCA (\%)} & OBA ($\downarrow$) \\
      \midrule

      & \multicolumn{1}{l}{LayoutGPT~\cite{layoutgpt}} & 35.53 & 13.69 & 72.8 & 2913.6 && 32.80 & 8.99 & 67.6 & 2447.4 \\

      \midrule

      & \multicolumn{1}{l}{ATISS~\cite{atiss}} & 25.67 & 8.91 & 71.8 & 857.3 && 28.05 & 9.26 & 63.2 & 702.4 \\

      & \multicolumn{1}{l}{DiffuScene~\cite{diffuscene}} & 21.54 & 6.40 & 69.7 & 341.1 && 23.06 & 5.35 & 57.7 & 266.4 \\

      & \multicolumn{1}{l}{\textbf{DeBaRA (ours)}} & \textbf{18.89} & \textbf{3.57} & \textbf{68.3} & \textbf{167.8} && \textbf{22.04} & \textbf{4.41} & \textbf{52.4} & \textbf{132.8} \\

      \bottomrule
  \end{tabular}
  }
  \label{tab:arrangegen}
\end{table*}

\subsection{3D Scene Synthesis and Self Score Evaluation}

\begin{figure}[b]
  \centering
  \includegraphics[width=1.\textwidth]{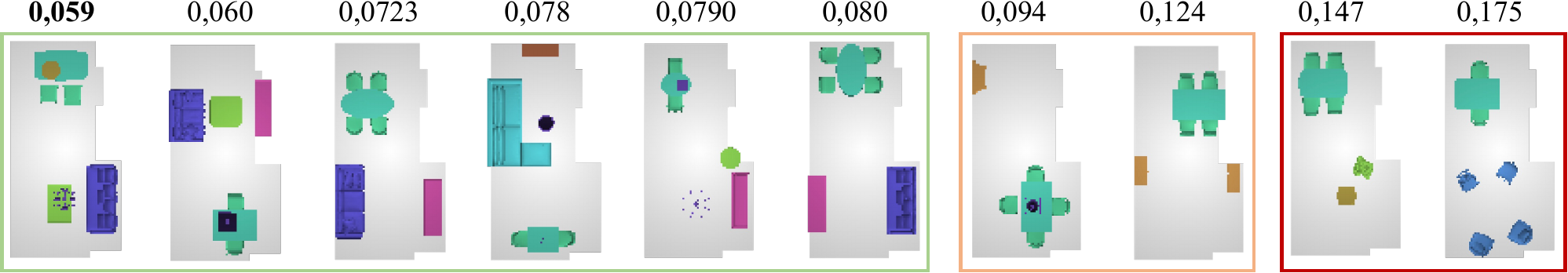}
  \caption{Top-down views of scenes generated by DeBaRA from several \textit{conditioning candidates} provided by a LLM and their associated SSE values. We qualitatively observe that lower scores (\textit{green}) corresponds to more natural layouts while higher scores (\textit{red}) can be filtered out.{\protect}}
  \label{fig:sse-results}
\end{figure}

We demonstrate competitive or state-of-the-art capabilities on 3D scene synthesis against methods that have been specifically trained for this task. We consider several settings depending on the source of input object categories and report our results in Table~\ref{tab:synthesis}. First, we observe that randomly picking input semantics from the training set (\textit{Dataset Random}) or taking the set \(\cc\) generated by \textit{LayoutGPT}~\cite{layoutgpt} outperform baselines by a significant margin on the 3D-FRONT living rooms test set. Then, to measure the individual impact of SSE, we compare a setup in which input semantics are selected from a set of LLM-generated ones, either randomly (\textit{LLM}) or by applying \textit{SSE}. As LLMs often hallucinate or produce out-of-distribution sets, our procedure consistently improves realism and validity of the synthesized indoor scenes, which can also be qualitatively observed in Figure~\ref{fig:sse-results}. These results further validate our choice to focus solely on 3D spatial features of objects.

\begin{table*}[htbp]
  \caption{Quantitative experiment results on \textbf{3D scene synthesis}. DeBaRA is evaluated in various settings based on the source of object semantic categories \(\cc\). Precise settings are detailed and discussed in Appendix~\ref{sec:llmprompting}. DeBaRA outperforms established baselines on most evaluation metrics.{\protect}}
  \resizebox{\textwidth}{!}{%
  \begin{tabular}{l c cccc c cccc c}
  \toprule

   \multicolumn{3}{c}{\multirow{2.5}{*}{\textbf{Methods}}} & \multicolumn{4}{c}{\textbf{Living Rooms}} && \multicolumn{4}{c}{\textbf{Dining Rooms}}\\

  \cmidrule(lr){4-7} \cmidrule(lr){9-12}

  & \multicolumn{2}{c}{} & \makecell{FID ($\downarrow$)} & KID ($\downarrow$) & SCA (\%) & OBA ($\downarrow$) && \makecell{FID ($\downarrow$)} & KID ($\downarrow$) & SCA (\%) & OBA ($\downarrow$)\\

  \midrule

  \multicolumn{3}{c}{LayoutGPT~\cite{layoutgpt}} & 34.26 & 10.17 & 72.1 & 2902.7 && 37.78 & 11.31 & 60.2 & 1982.1 \\

  \multicolumn{3}{c}{ATISS~\cite{atiss}} & 27.02 & 10.99 & 73.0 & 848.4 && 28.26 & 9.28 & 58.2 & 759.1 \\

   \multicolumn{3}{c}{DiffuScene~\cite{diffuscene}} & 21.64 & 5.94 & \textbf{66.0} & 323.1 && \textbf{23.85} & 5.66 & 54.6 & 289.8 \\

  \midrule

  \multicolumn{1}{c}{\multirow{2}{*}{\textbf{DeBaRA}}} & \multicolumn{2}{l}{\textit{LayoutGPT}} & 20.97 & 3.53 & 69.8 & 193.0 && 26.67 & 7.14 & 56.6 & 151.8 \\

  & \multicolumn{2}{l}{\textit{Dataset Random}} & \textbf{19.52} & \textbf{3.53} & 67.6 & \textbf{159.0} && 25.45 & \textbf{5.11} & \textbf{52.5} & \textbf{139.5} \\

  \midrule
  
  \midrule

  \multicolumn{1}{c}{\multirow{2}{*}{\textbf{DeBaRA}}} 

  & \multicolumn{2}{l}{\textit{LLM}} & 21.58 & 3.53 & 72.4 & 154.3 && 27.09 & 7.38 & 60.5 & 140.4 \\

  & \multicolumn{2}{l}{\textit{LLM + \textbf{SSE}}} & \textbf{20.59} & \textbf{3.47} & \textbf{70.7} & \textbf{152.0} && \textbf{24.50} & \textbf{5.34} & \textbf{54.0} & \textbf{134.4} \\

  \bottomrule
  \end{tabular}
  }
  \label{tab:synthesis}
\end{table*}

\subsection{Other Application Scenarios}

\begin{figure}[h]
  \begin{minipage}[h]{0.48\linewidth}
    \centering
    \begin{subfigure}[b]{0.49\textwidth}
      \centering
      \includegraphics[width=0.77\textwidth]{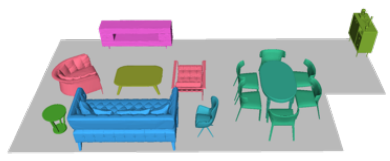}
      \caption{\small Ground Truth}
    \end{subfigure}
    \hfill
    \begin{subfigure}[b]{0.49\textwidth}
      \centering
      \includegraphics[width=0.77\textwidth]{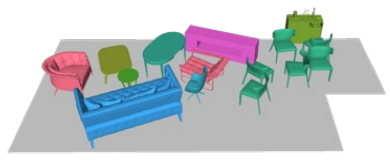}
      \caption{\small Noisy}
    \end{subfigure}
    \begin{subfigure}[b]{0.49\textwidth}
      \centering
      \includegraphics[width=0.77\textwidth]{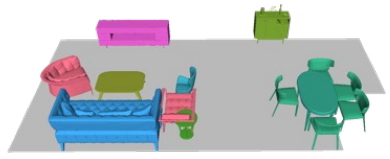}
      \caption{\small LEGO-Net~\cite{legonet}}
    \end{subfigure}
    \hfill
    \begin{subfigure}[b]{0.49\textwidth}
      \centering
      \includegraphics[width=0.77\textwidth]{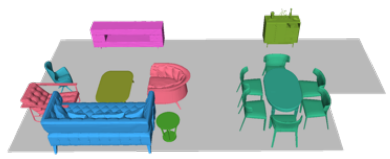}
      \caption{\small\textbf{DeBaRA}}
    \end{subfigure}
    \captionof{figure}{Qualitative comparison against LEGO-Net~\cite{legonet} on \textbf{scene re-arrangement}.}
    \label{fig:lego}
  \end{minipage}\hfill
  \begin{minipage}[h]{0.48\linewidth}
    \centering
    \captionof{table}{Quantitative evaluation on \textbf{scene re-arrangement}. DeBaRA is able to recover more realistic arrangements, closer to their initial \textit{noisy} configurations.}
    \resizebox{1.\textwidth}{!}{%
    \begin{tabular}{c c c c}
      \toprule
      
      \textbf{Method} & FID ($\downarrow$) & KID ($\downarrow$) & \makecell{Distance \\ Moved ($\downarrow$)} \\
      
      \midrule

      \vspace{1mm}
      
      \makecell{ LEGO-Net~\cite{legonet} \\ \textit{grad w/o noise} } & 26.81 & 13.18 & 0.094 \\
      
      \textbf{DeBaRA} & \textbf{24.92} & \textbf{9.47} & \textbf{0.082} \\
      
      \bottomrule
      \end{tabular}
      }
    \label{tab:lego}
  \end{minipage}
\end{figure}

We present DeBaRA's capabilities at performing additional controllable tasks. Notably, we include quantitative (Table~\ref{tab:lego}) and qualitative (Figure~\ref{fig:lego}) experimental evaluations against LEGO-Net~\cite{legonet} on scene-rearrangement. Results highlight that our method is able to recover more realistic arrangements, while being closer to their initial, \textit{messy} configurations. This is remarkable as the LEGO-Net baseline has been specifically trained to perform this task.

We also provide additional re-arrangement results and showcase DeBaRA's scene completion capabilities, by inserting objects from a list of additional semantics, in Figure~\ref{fig:arrange-complete}.

\subsection{Ablations}

We evaluate the individual contributions of some of our framework's key components on the base 3D layout generation task. Notably, results reported in Table~\ref{tab:ablation-training} highlight the advantage of our novel objective (Section~\ref{section:method:objective}) over common formulations as well as the benefits of modeling both the unconditional and class-conditional densities of 3D layouts during training (Section~\ref{section:method-framework}).

\begin{table*}[!h]
  \centering
  \caption{Ablation study on DeBaRA \textbf{training setup}. We evaluate the individual impact, on 3D layout generation, of different learning objectives \(\mathcal{L}\) and of applying conditioning dropout with rate \(\pdrop\). Notably, the use of our novel Chamfer distance results in a significant performance increase.
  }
  \resizebox{\textwidth}{!}{%
  \begin{tabular}{ccc c cccc c cccc}
  \toprule
      & \multicolumn{2}{c}{\textbf{Ablation Setting}} && \multicolumn{4}{c}{\textbf{Living Rooms}} && \multicolumn{4}{c}{\textbf{Dining Rooms}}\\

      \cmidrule(lr){2-3} \cmidrule(lr){5-8} \cmidrule(lr){10-13}

      & \(\mathcal{L}(\hat{\mathcal{O}},\mathcal{O})\) & \(\pdrop\) && \multicolumn{1}{c}{FID ($\downarrow$)} & \makecell{KID ($\downarrow$)} & {SCA (\%)} & OBA ($\downarrow$) && \makecell{FID ($\downarrow$)} & KID ($\downarrow$) & {SCA (\%)} & OBA ($\downarrow$) \\
      \midrule

      & \(MSE\) & 0.0 && 21.66 & 6.55 & 70.9 & 237.0 && 23.89 & 5.51 & 56.9 & 136.5 \\

      & \(CD\) \textit{standard} & 0.0 && 21.76 & 7.05 & 71.7 & 225.1 && 25.21 & 6.75 & 59.4 & 294.7 \\

      & \(CD\) \textit{semantic-aware} \textbf{(ours)} & 0.0 && 19.89 & 4.82 & \textbf{63.5} & 220.0 && 22.60 & 4.87 & 53.4 & 159.4 \\

      \midrule

      & \(CD\) \textit{semantic-aware} \textbf{(ours)} & 0.2 && \textbf{18.89} & \textbf{3.57} & 68.3 & \textbf{167.8} && \textbf{22.04} & \textbf{4.41} & \textbf{52.4} & \textbf{132.8} \\

      \bottomrule
  \end{tabular}
  }
  \label{tab:ablation-training}
\end{table*}

\subsection{Additional Results}

\paragraph{Complex Floor Plans}

\begin{figure}[H]
  \centering
  \includegraphics[width=1.\textwidth]{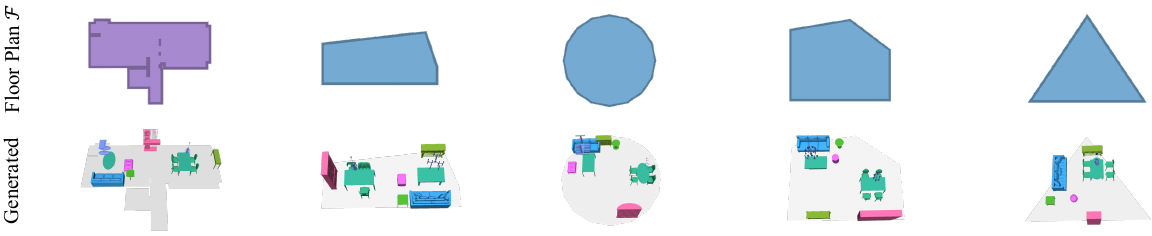}
  \caption{Generated layouts from a given set of objects and \textbf{complex floor plans}, \textcolor{purplefp}{selected} from the 3D-FRONT test set or \textcolor{bluefp}{handcrafted} to irregular, out-of-distribution shapes. While challenging, DeBaRA is able to output plausible layouts in which objects are scattered across the input floor plans.}  
  \label{fig:ood-fp}
\end{figure}

We notice that the 3D-FRONT dataset mostly contains \textit{simple} floor maps (i.e., single room, squared, rectangular) both for training and evaluation. As a result, we manually designed irregular floor shapes and report DeBaRA's generation in Figure~\ref{fig:ood-fp}, which further highlights the robustness of our method and its consideration for the conditioning input.

\paragraph{Iterative Sampling}

\begin{figure}[H]
  \centering
  \includegraphics[width=1.\textwidth]{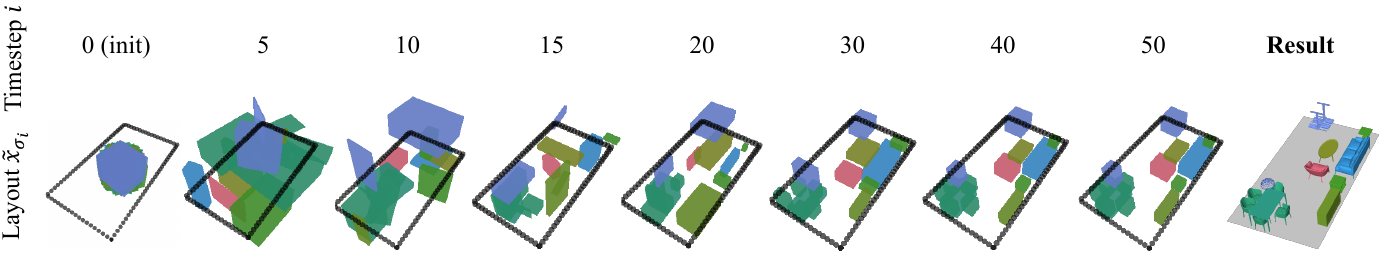}
  \caption{Visualization of intermediate layouts throughout the DeBaRA \textbf{denoising process}. \textit{Coarse} object attributes (positions, rotations and dimensions) are determined in the early steps, and then refined in the late iterations.}  
  \label{fig:iterative-denoising-process}
\end{figure}

We provide a visualization of the iterative denoising process over time when generating a 3D layout from arbitrarily initialized object bounding boxes in Figure~\ref{fig:iterative-denoising-process}.

\paragraph{Generation Variety and Validity}

\begin{figure}[htbp]
  \begin{minipage}[b]{0.43\linewidth}
    \centering
    \includegraphics[width=0.70\textwidth]{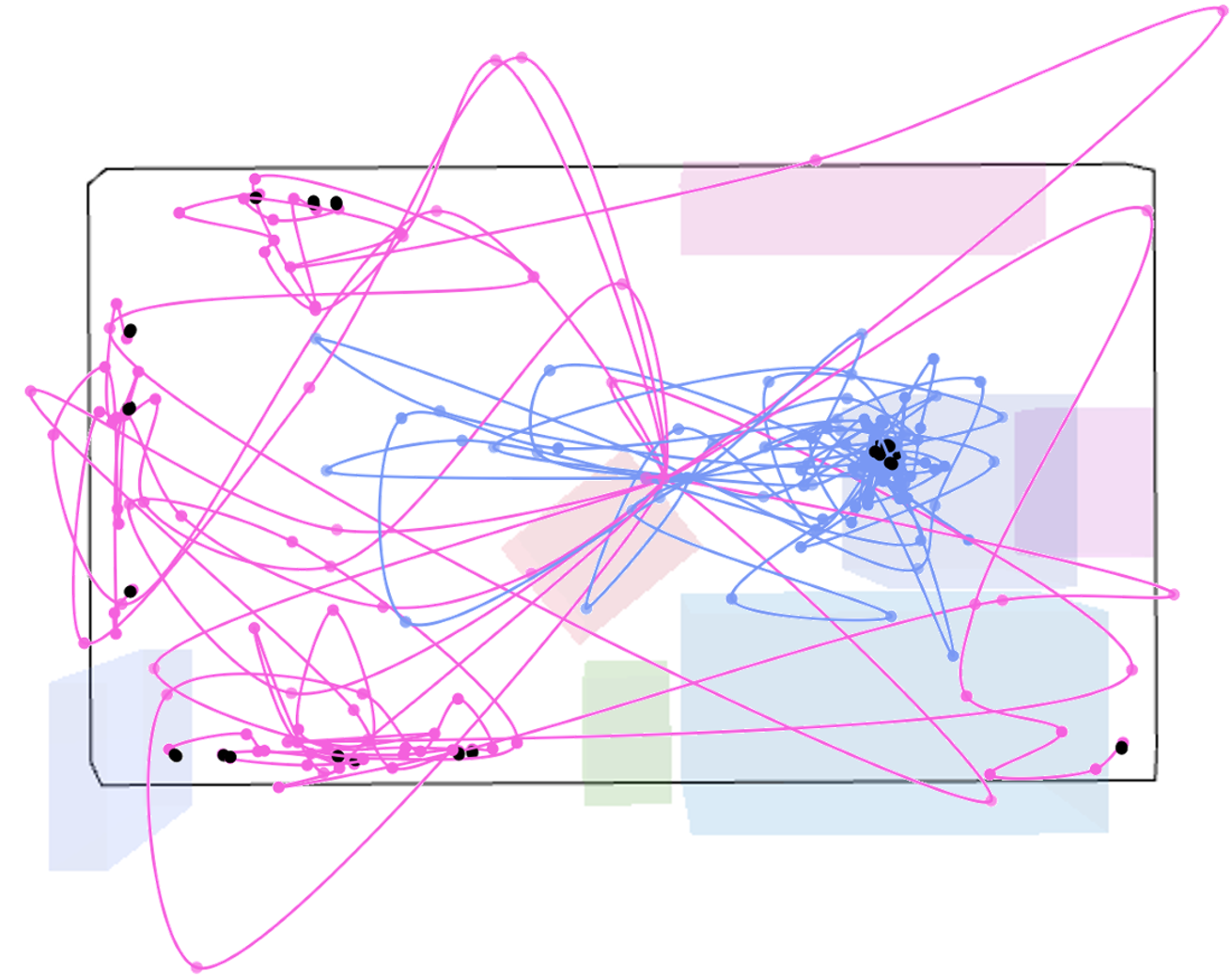}
    \captionof{figure}{Visualization of sampling trajectories and final positions of a pair of objects, inserted into a scene across ten trials.}
    \label{fig:denoising-process}
  \end{minipage}\hfill
  \begin{minipage}[b]{0.56\linewidth}
    \centering
    \captionof{table}{Generation times are averaged on the 3D-FRONT~\cite{3dfront} living room test subset. DeBaRA is implemented with \(T=50\) sampling steps and \(\Tsse=100\) Self Score Evaluation trials.}
    \resizebox{0.9\textwidth}{!}{%
      \begin{tabular}{c c c}
      \toprule
      
      \textbf{Method} & \makecell{ Network \\ Parameters ($10^6$) } & \makecell{ Generation \\ Time (s) } \\
      
      \midrule
      
      ATISS~\cite{atiss} & 36.1 & 0.160 \\
      
      DiffuScene~\cite{diffuscene} & 89.7 & 32.796 \\
      
      \textbf{DeBaRA} & 12.2 & 0.488 \\

      \textbf{DeBaRA + SSE} & 12.2 & 0.894 \\
      
      \bottomrule
      \end{tabular}
      }
    \vspace{4pt}
    \label{tab:efficiency}
  \end{minipage}
\end{figure}

We also perform scene completion by adding a \textcolor{pinkobject}{bookshelf} and a \textcolor{blueobject}{coffee table}, repeat the experiment ten times and report in Figure~\ref{fig:denoising-process} the denoising object trajectories, intermediate and final positions (colored and black dots respectively). This allows to observe the variety of predicted layouts. Notably, we can see that the \textcolor{pinkobject}{bookshelf} ends up in various different positions, always next to a wall, while the \textcolor{blueobject}{coffee table}, which is functionally constrained by its surrounding objects, is placed at similar valid locations.

\paragraph{Network Efficiency}

Finally, we compare the number of parameters as well as the sampling (i.e., generation) time, measured on the 3D layout generation task, of our DeBaRA backbone with those of other recent data-driven approaches in Table~\ref{tab:efficiency}. We can see that our lightweight architecture is bridging the gap with autoregressive methods in terms of inference efficiency.

\section{Conclusion, Limitations and Future Work}
\label{sec:conclusion}

In this paper we proposed DeBaRA, a novel score-based framework, which achieves state-of-the-art results in 3D layout generation. Our approach is distinctive in its design choices, which both favor data-efficiency with enhanced spatial reasoning, while, at the same time, enabling a range of applications such as scene re-arrangement and completion. Furthermore, we introduce a novel Self Score Evaluation procedure, which allows us to leverage a trained model to select the conditioning signals, which lead to the most plausible results. Overall, our work is the first to unify the conditioning and prediction spaces of score-based models within the context of 3D generative layout.

While powerful, our method currently does not enforce physical constraints between interacting objects, which can lead to collisions. We also assume that object semantic classes are selected among a finite set of predefined categories. Finally, we do not enforce \textit{style consistency} between objects, which can, nevertheless, be performed at retrieval time.

We believe that our approach can enhance other generative models (e.g., architectural layouts, images) by both evaluating the quality and by promoting more plausible 3D layout designs. Furthermore, it will be interesting to combine our approach with encoders from other modalities for a unified multi-modal layout generation.

\newpage

\begin{ack}
  We thank Despoina Paschalidou for the preprocessing code base and Jiapeng Tang for the mesh rendering pipeline. We also acknowledge the anonymous reviewers for their insightful suggestions. This work was supported by Dassault Systèmes SE. The views and conclusions contained in the paper are those of the authors and should not be interpreted as representing official policies, either expressed or implied, of the company.
\end{ack}

{
    \small
    \bibliographystyle{plain}
    \bibliography{main}
}


\appendix
\newpage

\section*{Appendix}

\section{Score-based Framework}
\label{sec:sup:framework}

In this section, we give additional details on the score-based parameterization that we adopt to learn the distribution of 3D layouts \(p_{\stext{data}}(\xx)\) and the sampling strategy used to generate new samples from the resulting trained denoiser \(D_\theta\).

\subsection{Training}

Score-based approaches model the score (i.e., the gradient of log-probability density w.r.t. the data) of marginal distributions \(p_\sigma(\xx)\) obtained by perturbing the data with Gaussian noise \(\boldsymbol{\epsilon} \sim \mathcal{N}(\boldsymbol{0},\boldsymbol{\mathbf{I}})\) at magnitudes \(\sigma\). In practice, the score can be effectively approximated by a noise-conditioned denoiser that outputs clean samples from noisy ones, then, \(\nabla_{\xx} \log p_\sigma(\xx) =  (D_\theta(\xx;\yy,\sigma) - \xx)/\sigma^2\).

Parameterizing the denoiser to output \(\xx\) from its corrupted version directly is not ideal as the input magnitude varies greatly depending on the current noise level. Instead, Karras et al.~\cite{karras} propose in their EDM diffusion framework a preconditioning of the denoiser whose output is now derived from a trainable network \(F_\theta\) that either predicts the clean signal \(\xx\), the noise \(\boldsymbol{\epsilon}\) or something in between, depending on the value of \(\sigma\). More formally, it can be expressed as:
\begin{equation}
  D_\theta(\xx;\yy,\sigma) = c_{\stext{skip}}(\sigma)\,\xx + c_{\stext{out}}(\sigma)\,F_\theta\bigl(c_{\stext{in}}(\sigma)\, \xx;\yy,c_{\stext{noise}}(\sigma)\bigr)
\end{equation} 
The preconditioning function \(c_{\stext{skip}}\) amplifies the network error as little as possible while \(c_{\stext{in}}\) and \(c_{\stext{out}}\) scale respectively the input and output to have unit variance. Following~\cite{karras}, we set:
\begin{equation}
  c_{\stext{skip}} = \frac{\sigma^{2}_{\stext{data}}}{\sigma^{2}_{\stext{data}} + \sigma^{2}}\ ; \ c_{\stext{in}}=\frac{1}{\sqrt{\sigma^{2}_{\stext{data}} + \sigma^{2}}} \ ; \ c_{\stext{out}} = \frac{\sigma \cdot \sigma_{\stext{data}}}{\sqrt{\sigma^{2}_{\stext{data}} + \sigma^{2}}} \ ; \ c_{\stext{noise}} = \frac{\ln(\sigma)}{4}
\end{equation}
Note that in our case, the value of \(\sigma_{\stext{data}}\) should preferably be computed channel-wise for each attribute of \(\xx = (\pp, \rr, \dd) \in \mathbb{R}^{8}\), as object positions, rotations and dimensions typically have different standard deviations. In practice, we compute \(\sigma_{\stext{data}}^{\pp}\) and \(\sigma_{\stext{data}}^{\dd}\) from the training data and arbitrarily set \(\sigma_{\stext{data}}^{\rr}\ = (0.5, 0.5)\). During training, noise values \(\sigma\) are drawn from a centered normal distribution of variance \(0.25\), which concentrates training on \textit{medium} noise levels. Spatial values \(\pp\) and \(\dd\) of each training layout \(\xx\) are normalized based on the maximum extent of the scene's floor plan \(\mathcal{F}\), which ensures that all the network's inputs and outputs are scaled in \([-1, 1]\). To model both the class-conditional and unconditional layout densities, we perform conditioning dropout on object categories \(\cc\) with a rate \(\pdrop = 0.2\).

Finally, the training objective can be expressed by introducing our semantic-aware Chamfer reconstruction loss following Equation~\ref{eq:diffusion-loss}. As in EDM, we use \(\lambda(\sigma) = 1 / c_{\stext{out}}^2 \) to get a uniform weighting across noise levels.

\subsection{Sampling}

At test time, the reverse SDE~\cite{karras,song2020score} associated to the continuous-time diffusion process is used to generate novel samples from a standard normal distribution using numerical solvers. It depends on the score approximated during training:
\begin{equation}
  d\xx = \bigl[f(\xx,t) - g(t)^2 \nabla_{\xx} \log p_t(\xx)\bigr]dt + g(t)dw
\end{equation}
where \(f(\cdot, t) : \mathbb{R}^d \rightarrow \mathbb{R}^d\) and \(g(\cdot) : \mathbb{R} \rightarrow \mathbb{R}\) are respectively the drift and diffusion coefficients and \(w\) is the standard stochastic Wiener process.

In practice, we use EDM~\cite{karras} 2\textsuperscript{nd} order Runge-Kutta stochastic sampler (see Algorithm~\ref{alg:sampler}), that resembles the predictor/corrector framework from Song et al.~\cite{song2020score} and provides a good trade-off between generation quality and number of function evaluations (NFE).

\begin{algorithm}[t]
  \caption{EDM Stochastic Sampler~\cite{karras}}
  \label{alg:sampler}
  \begin{algorithmic}[1]
      \Procedure{LayoutSampler}{$D_{\theta}(\xx;\yy,\sigma), \xx_0, t_{i \in \{0,\dots ,T-1\}}, \gamma_{i \in \{0,\dots ,T-1\}}, S_{\stext{noise}}$}
      \For{$i\in \{0,\dots ,T-1\}$ }
      \State \textbf{sample} $\boldsymbol{\epsilon}_i \sim \mathcal{N}(\boldsymbol{0},S_{\stext{noise}}^2 \textbf{I})$
      \State $\hat{t}_i \gets t_i + \gamma_i t_i$
      \State $\hat{\xx}_i \gets \xx_i + \sqrt{\hat{t}_i^2-t_i^2}\boldsymbol{\epsilon}_i$
      \State $\bg_i \gets \left(\hat{\xx}_i - D_{\theta}(\hat{\xx}_i;\yy,\hat{t}_i)\right)/\hat{t}_i$
      \State $\xx_{i+1} \gets \hat{\xx}_i +(t_{i+1}-\hat{t}_i)\bg_i $
      \If{$t_{i+1}\neq 0$}
      \State $\bg_i' \gets \left( \xx_{i+1} - D_{\theta}(\xx_{i+1};\yy,t_{i+1}) \right)/t_{i+1}$
      \State $\xx_{i+1} \gets \hat{\xx}_i + \frac{1}{2}(t_{i+1}-\hat{t}_i)(\bg_i +\bg_i')$
      \EndIf
      \EndFor
      \EndProcedure
  \end{algorithmic}
\end{algorithm}

It is based on a \(T\)-step discretization of the reverse SDE~\cite{karras,song2020score}, with timesteps \(t_{i \in \{0,\dots ,T-1\}}\) decreasing from \(\sigmamax\) (\(i=0\)) to \(\sigmamin\) (\(i=T-1\)). Following~\cite{karras}, we use:
\begin{equation}
  t_{i<T} = \left( {\sigmamax}^{\frac{1}{\rho}} + \frac{i}{T-1}\left({\sigmamin}^{\frac{1}{\rho}} - {\sigmamax}^{\frac{1}{\rho}}\right)\right)^{\rho}, \hspace*{3mm} t_T=0
\end{equation}
Note that \(t\) and noise level \(\sigma\) can be used interchangeably. The \(\rho\) parameter is tuned to dedicate more steps of the denoising process to smaller or larger noise levels. The \(\sigmamin\) value should be small enough so that the model estimates the best approximation of the score and sample a precise layout. On the other hand, \(\sigmamax\) should be large enough to sample various layouts. The amount of \textit{fresh} noise injected at the beginning of each denoising step is defined by \(\gamma_{i \in \{0,\dots ,N-1\}}\). Similar to Wei et al.~\cite{legonet} and Karras et al.~\cite{karras}, we qualitatively observed that adding noise in the final timesteps, i.e., when the layout is close to its final configuration, leads to less precise results. As a result, an additional \(\Smin\) parameter is set so that \(\gamma_i = 0\) when \(t_i < \Smin\).

For 3D layout generation, we use \(T=50\) timesteps and set \(\sigmamax = 1.0\), \(\sigmamin = 0.005\), \(\rho = 7\), and \(\Smin = 0.005\). Note that, although we perform conditioning dropout during training, we didn't find the need to amplify the strength of the input categories using any classifier-free guidance~\cite{cfg} scale at sampling time.

Additionally, Table~\ref{tab:ablation-sampling} shows that sampling from DeBaRA using EDM~\cite{karras} 2\textsuperscript{nd} order stochastic procedure (Algorithm~\ref{alg:sampler}) outperforms ancestral DDPM~\cite{ddpm} sampling using a fraction of the denoising steps. This, combined with our lightweight architecture, enables \textit{real-time} (<1s) generation.

\begin{table*}[!h]
  \caption{Ablation study on DeBaRA \textbf{sampling strategy}. Metrics are computed on the 3D layout generation task.}
  \centering
  \resizebox{\textwidth}{!}{%
  \begin{tabular}{ccc c ccccc c ccccc}
  \toprule
      & \multicolumn{2}{c}{\textbf{Sampler}} && \multicolumn{5}{c}{\textbf{Living Rooms}} && \multicolumn{5}{c}{\textbf{Dining Rooms}}\\

      \cmidrule(lr){2-3} \cmidrule(lr){5-9} \cmidrule(lr){11-15}

      & Alg. & Steps && \multicolumn{1}{c}{FID ($\downarrow$)} & \makecell{KID ($\downarrow$)} & {SCA (\%)} & OBA ($\downarrow$) & Time (s) && \makecell{FID ($\downarrow$)} & KID ($\downarrow$) & {SCA (\%)} & OBA ($\downarrow$) & Time (s) \\
      \midrule

      & DDPM & 1000 && 21.12 & 5.65 & \textbf{67.4} & 268.9 & 5.144 && 23.18 & 5.78 & 53.3 & 202.9 & 4.925 \\

      & EDM & 25 && 19.53 & 3.95 & 69.4 & \textbf{159.5} & \textbf{0.247} && \textbf{21.95} & \textbf{4.26} & 54.7 & 140.5 & \textbf{0.248} \\

      & EDM & 50 && \textbf{18.89} & \textbf{3.57} & 68.3 & 167.8 & 0.488 && 22.04 & 4.41 & \textbf{52.4} & \textbf{132.8} & 0.514 \\

      \bottomrule
  \end{tabular}
  }
  \label{tab:ablation-sampling}
\end{table*}

\section{Implementation}
\label{sec:sup:imp}

We provide in this section additional implementation details on our model architecture and training configurations, illustrated in Figure~\ref{fig:archi}. We also detail how baselines have been retrained and used at test time to ensure a fair and relevant comparison with our approach.

\subsection{Network Architecture}
\label{sec:sup:imp:network}

\paragraph{Shared Object Encoder}

The shared object encoder embeds each object \(o_i\) from its input 3D spatial values \(\xx_i\) and semantic category \(\cc_i\). Triplets of scalar values of the object's position \(\pp_i\) and dimension \(\dd_i\) are encoded with fixed sinusoidal positional encoding of \(32\) frequencies following~\cite{legonet}:
\begin{equation}
  PE(s) = \bigl\{\sin(128^{j/31}s), \cos(128^{j/31}s)\bigr\}_{j=0}^{31} \in \mathbb{R}^{64}
\end{equation}
Applying this module projects \(\pp_i\) and \(\rr_i\) in \(\mathbb{R}^{192}\). It is similarly applied to \(\rr_i = (\cos(\theta_i), \sin(\theta_i))\) to get a feature in \(\mathbb{R}^{128}\), that is additionally fed to a linear layer to obtain a \(192\)-dimensional attribute.

The object semantic class \(\cc_i\), represented as a one-hot vector among \(k\) classes is encoded in \(\mathbb{R}^{192}\) by an MLP with \(2\) a hidden layer of 128 units and LeakyReLU activation. Respective object spatial and semantic encodings are then concatenated to form an object token \(\mathcal{T}_{o_i}\) of dimension \(4 \times 192 = 768\).

\paragraph{Floor Encoder}

The scene's conditioning floor plan \(\mathcal{F}\) is embedded by a PointNet~\cite{qi2017pointnet} module, similar to~\cite{legonet}. To do so, we first extract the floor's 2D polygon using the output of ATISS~\cite{atiss} preprocessing and sample \(P=100\) evenly spaced points on its contour. The PointNet backbone\footnote{\url{https://github.com/fxia22/pointnet.pytorch}} produces a \(1024\)-dimensional feature, that is further passed to a linear layer to get the appropriate floor token \(\mathcal{T}_{\mathcal{F}} \in \mathbb{R}^{768}\).

\paragraph{Noise Level Encoder}

We encode the noise level \(\sigma\) as a token \(\mathcal{T}_\sigma \in \mathbb{R}^{768}\) obtained by subsequently applying \(PE(\sigma)\) and a linear layer with LeakyReLU activation.

\paragraph{Transformer Encoder}

Our transformer encoder that computes new representations \(\mathcal{\hat{T}}\)  is composed of multi-head self-attention and feedforward layers, following the original paper~\cite{vaswani2017attention} and implementation from the PyTorch~\cite{paszke2019pytorch} API. Importantly, we don't enforce ordering of any input token and pass an additional padding mask to handle sequences of different lengths. We stack \(3\) encoder layers, each having \(4\) attention heads and a feedforward hidden dimension of \(512\).

\paragraph{Shared Object Decoder}

The final shared object decoder produces the network's predicted spatial values \(\hat{\xx_i} \in \mathbb{R}^8\) for each of the \(N\) objects from their respective \(\mathcal{\hat{T}}_{o_i}\) embeddings. It is implemented as an MLP with three layers of \(512\), \(128\), and \(8\) units, using LeakyReLU activations and a dropout rate of \(0.1\).

\subsection{Training Protocol}

During training, the network is optimized towards our semantic-aware Chamfer loss, that can be efficiently implemented with appropriate broadcasting. We trained our models separately on the 3D-FRONT~\cite{3dfront} \textit{living room} and \textit{dining room} subsets for \(3000\) epochs, with a batch size of \(32\) and monitor the validation loss to avoid overfitting of the training set in the late iterations. We use the AdamW~\cite{loshchilov2018decoupled} optimizer with its PyTorch default parameters and learning rate \(\eta=10^{-4}\), scheduled with a linear warmup phase for the first \(50\) epochs, starting at \(\eta \times 0.01\). Following this, a cosine annealing schedule~\cite{loshchilov2022sgdr} reduces \(\eta\) to a minimum of \(10^{-8}\) over \(2200\) epochs. Finally, we randomly apply rotations to the training scenes as data augmentation.

\subsection{Baselines}
\label{sec:sup:baselines}

\paragraph{ATISS}

ATISS~\cite{atiss} is an autoregressive, permutation-invariant transformer that treats 3D scene synthesis as an unordered set generation task. The model is natively conditioned on the room's floor plan, from which it extracts features using a ResNet-18~\cite{he2016deep} applied on a top-down binary projection. The model predicts the semantic class, location, rotation and dimension of the next object to be inserted to the current layout configuration. As our method, it also supports inserting objects from their semantic categories given as input, which is the setting that we used to report experimental results on the 3D layout generation task (see Table~\ref{tab:arrangegen} and Figure~\ref{fig:arrangegen}). We retrained the model on each 3D-FRONT subset using the authors' implementation.\footnote{\url{https://github.com/nv-tlabs/ATISS}}

\paragraph{DiffuScene}

DiffuScene~\cite{diffuscene} employs a DDPM to perform 3D scene synthesis, by learning to denoise unordered sets of objects that are each represented by all their attributes, i.e., location, size, orientation, semantic category and \textit{shape code}. Although the paper specifically mentions not being conditioned on the room's bounds, we found out the official implementation supports this feature that we enabled to retrain the model on the 3D-FRONT subsets, with other settings set to those of the authors. In practice and similar to ATISS~\cite{atiss}, a ResNet-18 backbone is used to extract features from the floor plan's projection mask. The resulting encoding is passed to an MLP whose output is added to the diffusion timestep embedding, as in~\cite{imagen}. To assess the effectiveness of this conditioning mechanism and validate the relevance of this baseline, we report metrics obtained for the 3D layout generation task, using both a floor-conditioned and an unconditional (\textit{vanilla}) trained DiffuScene model on the \textit{living room} subset in Table~\ref{tab:diffuscenecond}. Note that, top-down view of scenes, that we use to evaluate e.g., FID and KID scores, feature the floor plan, which penalizes generated configurations that don't properly take it into account.

\begin{table*}[!h]
  \caption{Quantitative impact of DiffuScene~\cite{diffuscene} floor plan conditioning on 3D layout generation. Results indicate that the additional input has successfully been learned, resulting in a solid baseline.}
  \centering
  \resizebox{0.5\textwidth}{!}{%
  \begin{tabular}{l c c c}

    \toprule

    \textbf{Model} & \textbf{FID} ($\downarrow$) &  \textbf{KID} ($\downarrow$) & \textbf{OBA} ($m^2$)\\

    \midrule

    DiffuScene \textit{vanilla} & 41.30 & 22.92 & 1621.5 \\

    DiffuScene \textit{floor} & \textbf{21.54} & \textbf{6.40} & \textbf{341.1} \\

    \bottomrule
  \end{tabular}
  }
  \label{tab:diffuscenecond}
\end{table*}
      
To perform 3D layout generation from input semantic categories using DiffuScene (Table~\ref{tab:arrangegen}, Figure~\ref{fig:arrangegen}), we implemented DDPM inpainting of the object spatial features from their categories as an additional method within the official implementation\footnote{\url{https://github.com/tangjiapeng/DiffuScene}} and using default sampling settings with \(1000\) timesteps.

\paragraph{LayoutGPT}

LayoutGPT~\cite{layoutgpt} is a training-free approach that utilizes Large Language Models to generate layouts both in the image and the 3D scene domains, demonstrating competitive performance with learning-based approaches on 3D scene synthesis. To do so, the method consists in prompting a LLM with specific instructions and by adding \textit{supporting examples} from the training set, i.e., few-shot exemplars of expected, valid layouts. These examples are retrieved from the train set based on floor plan similarity computed from the binary masks with a test sample. We reimplemented LayoutGPT, using the official implementation\footnote{\url{https://github.com/weixi-feng/LayoutGPT}} for exact prompt and \textit{supporting examples} retrieval, but using a Meta Llama-3-8B\footnote{\url{https://huggingface.co/meta-llama/Meta-Llama-3-8B}} backbone instead of ChatGPT variants for local execution and better reproducibility.

To perform 3D layout generation, we include in the prompt the list of object semantic categories. Here is a typical LayoutGPT prompt for this task on the \textit{living room} subset:

\begin{figure}[h]
  \centering
  \includegraphics[width=1.\textwidth]{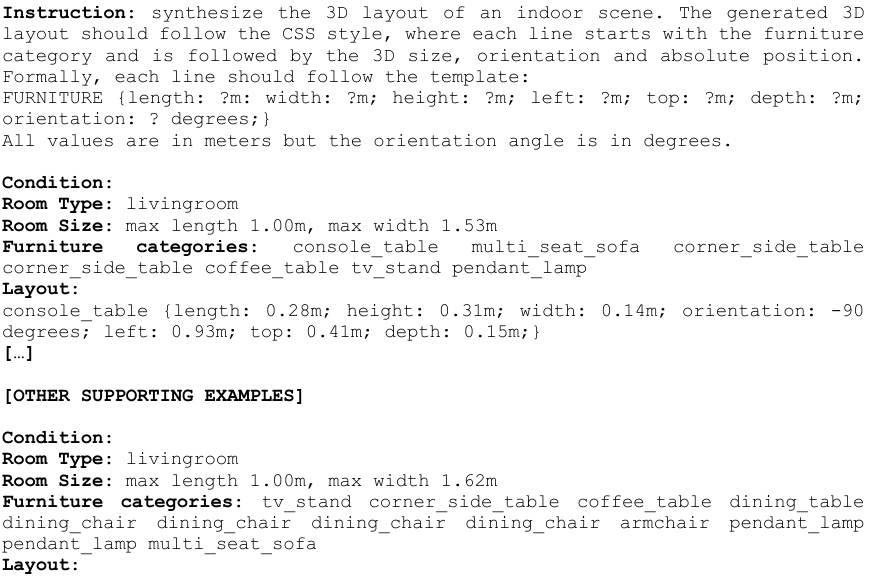}
\end{figure}
For 3D scene synthesis, we follow the paper and include the training set's object frequencies in the prompt. We set the LLM sampling temperature to \(0.7\) and maximum output tokens to \(1024\). We report LayoutGPT performance on this task in Table~\ref{tab:synthesis}, where we also  indicate metrics obtained by DeBaRA when using as conditioning input the same semantic set as the one generated by LayoutGPT for the corresponding test scene. In this setup, we largely outperform the baseline and even report state-of-the-art FID, KID and OBA scores on 3D-FRONT \textit{living room}.

\paragraph{LEGO-Net}

The LEGO-Net~\cite{legonet} model is specifically designed to perform 2D scene re-arrangement, i.e., recover a close \textit{clean} layout configuration from a \textit{messy}, perturbed one. It is trained using a regression loss on object position and rotation values, and proposes a Langevin dynamics-like iterative sampling procedure. In order to produce the results reported in Table~\ref{tab:lego}, we used authors' implementation,\footnote{\url{https://github.com/QiuhongAnnaWei/LEGO-Net}} in the \textit{grad without noise} setting (which is the best performing in the original paper), on the 3D-FRONT \textit{living room} test subset and with a scene perturbation level of \(0.25\).

\begin{figure}[t]
  \centering
  \includegraphics[width=1.\textwidth]{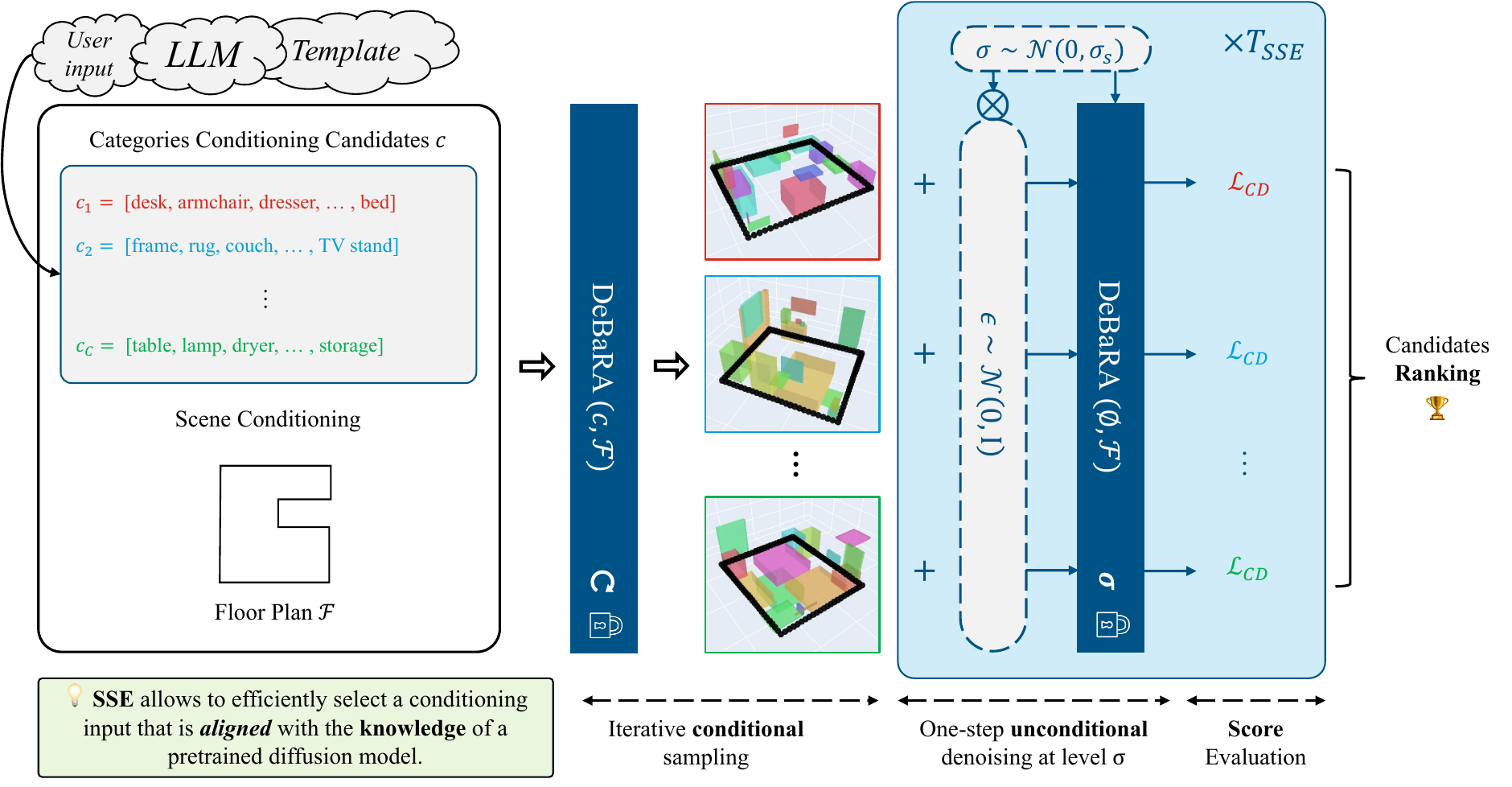}
  \caption{Self Score Evaluation (SSE) Pipeline. The method allows to leverage the knowledge of a model trained following our method to select valid sets of object categories.} 
  \label{fig:sse}
\end{figure}

\subsection{LLM Prompting}
\label{sec:llmprompting}

As mentioned in the main submission, we perform 3D scene synthesis using DeBaRA conditioned on LLM-generated sets of object semantics, that we optionally select via \textit{Self Score Evaluation} (Table~\ref{tab:synthesis}, Figure~\ref{fig:sse-results}). In practice, the generated categories have been obtained using Llama-3-8B~\cite{llama3} following the LayoutGPT~\cite{layoutgpt} prompting strategy. We experienced asking the language model to generate solely lists of object semantics using a few supporting examples and providing the dataset statistics, but we noticed that it was more prone to hallucinate and drift towards inconsistent generations than when generating complete layout configurations (i.e., including object position and orientation values).

The \textit{DeBaRA LLM} reported in Table~\ref{tab:synthesis} corresponds to the setup where we filter from the LLM-generated sets of categories those having the same number of objects as the considered test scene and randomly select one to condition DeBaRA. The \textit{DeBaRA LLM + SSE} setting is similar, but instead of picking a set randomly, the selection is performed by applying the SSE procedure.

\subsection{Computational Requirements}
\label{sec:sup:compute}

All the training and evaluation experiments as well as the computation of generation times reported in Table~\ref{tab:efficiency} have been performed on a single NVIDIA RTX A6000 GPU. When comparing our number of network parameters and generation times with those of ATISS~\cite{atiss} and DiffuScene~\cite{diffuscene}, we notice that DeBaRA is bridging the gap with autoregressive methods in terms of inference efficiency. This is made possible by our restricted output space that requires a more lightweight backbone as well as our choice of sampling procedure that leads to a favorable NFE / generation quality tradeoff.

\section{Self Score Evaluation}
\label{sec:sup:sse}

In this section, we provide additional content regarding our SSE procedure (Section~\ref{section:sse}), by illustrating it and by further assessing its expressive power.

\subsection{Pipeline}

The SSE formulation is expressed by Equations~\ref{eq:sse-candidates}~-~\ref{eq:sse-mce}. It follows the procedure outlined in Algorithm~\ref{alg:sse}. We additionally illustrate the SSE pipeline in Figure~\ref{fig:sse}. In our experiments, SSE is implemented using \(\Tsse=100\) trials, with noise levels \(\sigma\) drawn as in training.

\subsection{Additional Evaluation}

We additionally evaluate the expressive power of the SSE procedure in a \textit{toy} binary classification task: for each scene of the test set, we create a corrupted version by replacing a proportion \(p_\stext{rand}\) of the scene's object categories by random ones. We report the binary classification score obtained by SSE when asked to discriminate the corrupted set of semantics from the ground truth one. We perform the experiments \(10\) times to account for the inherent stochasticity of the experimental setup and report the results for several values of \(p_\stext{rand}\) in Table~\ref{tab:ssetoy}.

\begin{table*}[!ht]
  \caption{SSE performance at discriminating ground truth sets of categories from perturbed ones.}
  \resizebox{\textwidth}{!}{%
  \centering
\begin{tabular}{l c c c c c c}

  \toprule

  \textbf{Perturbation} & None &  Single & $p_\stext{rand} = 0.35$ & $p_\stext{rand} = 0.50$ & $p_\stext{rand} = 0.75$ & All \\

  \midrule

  \textbf{Accuracy (\%)} & \(50.6 \pm 4.39\) & \(71.7 \pm 2.20\) & \(73.8 \pm 3.38\) & \(78.5 \pm 1.84\) & \(80.9 \pm 2.62\) & \(84.0 \pm 2.90\) \\

  \bottomrule
\end{tabular}
}
\label{tab:ssetoy}
\end{table*}

We observe a significant gap in accuracy between the control experiment (none object is perturbed, meaning that the sets are equals) and the setting where only a single object has been swapped. It means that SSE is able to identify subtle misalignments between the network's knowledge and the provided conditioning candidates. 

\section{Additional Results}

In this section, we provide supplementary quantitative indicators measuring the validity of generated layouts with respect to the conditioning floor plan. We also showcase additional qualitative results and comparisons in multiple application scenarios.

\subsection{Bounding Metrics}
\label{sec:sup:res:bound}

In addition to the cumulated out of bounds objects area (OBA, in \(m^2\)) reported in Table~\ref{tab:arrangegen}, we indicate the rate of scenes having at least one object out of its bounds (OBR), and the cumulated number of out-of-bounds objects in the generated layouts (OBN) in Table~\ref{tab:arrangebounds}. For OBR and OBN, we consider an object to be out-of-bounds if at least \(20\%\) of its 2D bounding box surface is outside the floor's limits.

\begin{table*}[!h]
  \centering
  \caption{Quantitative bounding metrics on the 3D layout generation task. We observe that DeBaRA is consistently better at respecting the indoor floor plan by a significant margin.}
  \resizebox{\textwidth}{!}{%
  \begin{tabular}{l l ccc c ccc}
  \toprule
      & \multicolumn{1}{c}{\multirow{2.5}{*}{\textbf{Methods}}} & \multicolumn{3}{c}{\textbf{Living Rooms}} && \multicolumn{3}{c}{\textbf{Dining Rooms}}\\

      \cmidrule(lr){3-5} \cmidrule(lr){7-9}

      & & \multicolumn{1}{c}{OBA ($\downarrow$)} & \makecell{OBR ($\downarrow$)} & OBN ($\downarrow$) && \makecell{OBA ($\downarrow$)} & OBR ($\downarrow$) & OBN ($\downarrow$) \\
      \midrule

      & \multicolumn{1}{l}{LayoutGPT~\cite{layoutgpt}} & 2913.6 & 0.695 & 2119 && 2447.4 & 0.659 & 1720 \\

      \midrule

      & \multicolumn{1}{l}{ATISS~\cite{atiss}} & 857.3 & 0.744 & 1195 && 702.4 & 0.891 & 1603 \\

      & \multicolumn{1}{l}{DiffuScene~\cite{diffuscene}} & 341.1 & 0.652 & 742 && 266.4 & 0.628 & 640 \\

      & \multicolumn{1}{l}{\textbf{DeBaRA (ours)}} & \textbf{167.8} & \textbf{0.390} & \textbf{497} && \textbf{132.8} & \textbf{0.403} & \textbf{401} \\

      \bottomrule
  \end{tabular}
  }
  \label{tab:arrangebounds}
\end{table*}

\subsection{Qualitative Results}

\begin{figure}[H]
  \centering
  \begin{subfigure}[b]{0.25\textwidth}
    \centering
    \includegraphics[width=\textwidth]{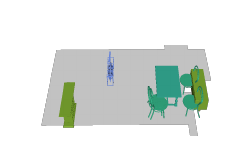}
    \includegraphics[width=\textwidth]{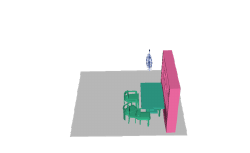}
    \includegraphics[width=\textwidth]{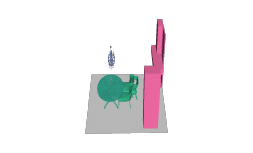}
    \includegraphics[width=\textwidth]{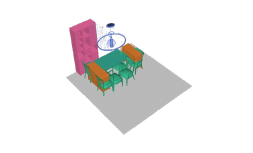}
    \includegraphics[width=\textwidth]{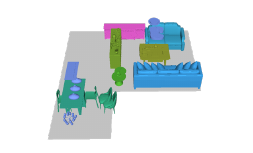}
    \includegraphics[width=\textwidth]{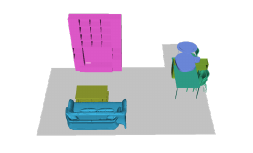}
    \includegraphics[width=\textwidth]{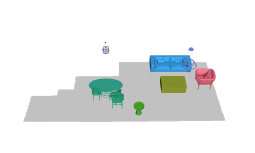}
    \includegraphics[width=\textwidth]{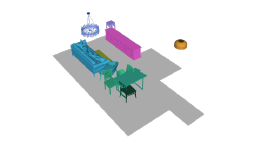}
    \includegraphics[width=\textwidth]{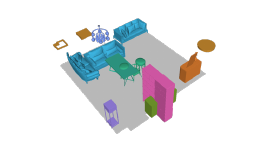}
    \caption{ATISS~\cite{atiss}}
  \end{subfigure}
  \hfill
  \vrule
  \hfill
  \begin{subfigure}[b]{0.25\textwidth}
    \centering
    \includegraphics[width=\textwidth]{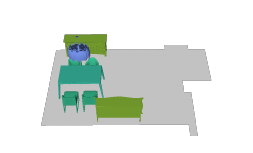}
    \includegraphics[width=\textwidth]{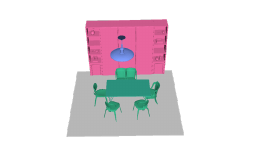}
    \includegraphics[width=\textwidth]{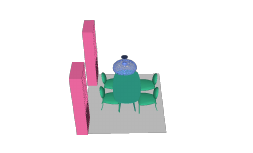}
    \includegraphics[width=\textwidth]{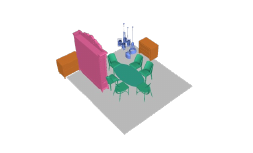}
    \includegraphics[width=\textwidth]{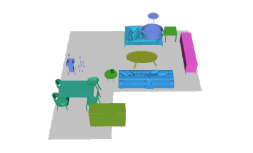}
    \includegraphics[width=\textwidth]{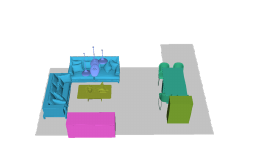}
    \includegraphics[width=\textwidth]{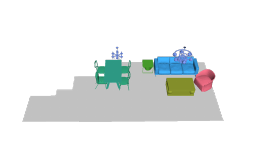}
    \includegraphics[width=\textwidth]{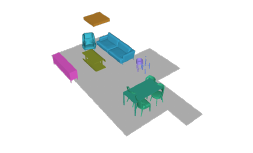}
    \includegraphics[width=\textwidth]{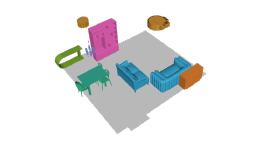}
    \caption{DiffuScene~\cite{diffuscene}}
  \end{subfigure}
  \hfill
  \vrule
  \hfill
  \begin{subfigure}[b]{0.25\textwidth}
    \centering
    \includegraphics[width=\textwidth]{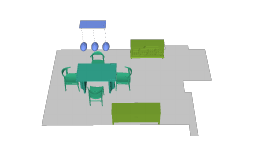}
    \includegraphics[width=\textwidth]{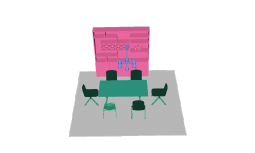}
    \includegraphics[width=\textwidth]{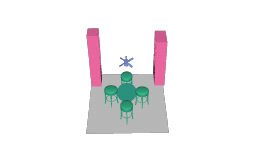}
    \includegraphics[width=\textwidth]{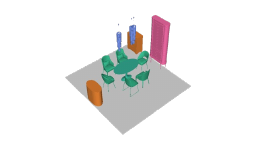}
    \includegraphics[width=\textwidth]{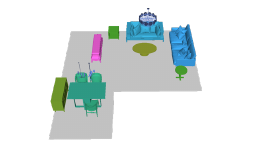}
    \includegraphics[width=\textwidth]{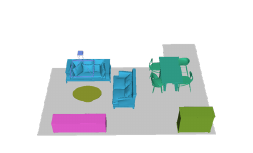}
    \includegraphics[width=\textwidth]{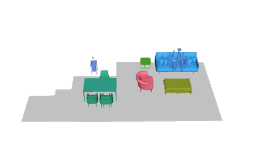}
    \includegraphics[width=\textwidth]{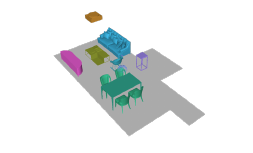}
    \includegraphics[width=\textwidth]{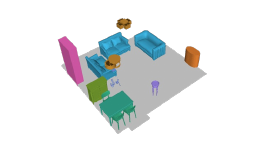}
    \caption{\textbf{DeBaRA}}
  \end{subfigure}
  \vspace{4pt}
  \caption{\textbf{Additional 3D layout generation results.} We compare our method by generating layouts from a list of object semantic categories and a floor plan. DeBaRA consistently produces more realistic arrangements while respecting the room's outline.}  
\end{figure}
\clearpage

\begin{figure}[H]
  \centering
  \begin{subfigure}[t]{0.19\textwidth}
    \centering
    \includegraphics[width=\textwidth]{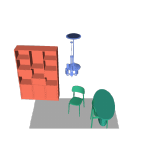}\vspace{-1pt}
    \includegraphics[width=\textwidth]{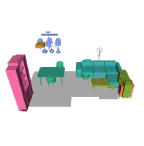}
    \includegraphics[width=\textwidth]{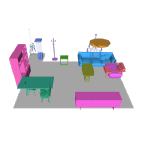}
    \includegraphics[width=\textwidth]{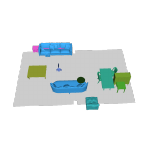}
    \includegraphics[width=\textwidth]{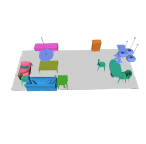}\vspace{-1pt}
    \includegraphics[width=\textwidth]{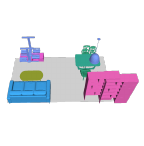}
    \includegraphics[width=\textwidth]{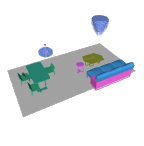}\vspace{3pt}
    \caption{ATISS~\cite{atiss}}
  \end{subfigure}
  \hfill
  \vrule
  \hfill
  \begin{subfigure}[t]{0.19\textwidth}
    \centering
    \includegraphics[width=\textwidth]{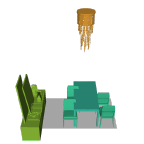}
    \includegraphics[width=\textwidth]{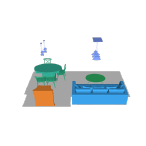}
    \includegraphics[width=\textwidth]{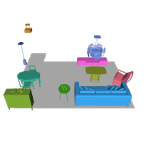}
    \includegraphics[width=\textwidth]{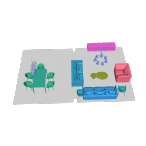}
    \includegraphics[width=\textwidth]{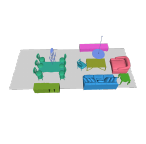}
    \includegraphics[width=\textwidth]{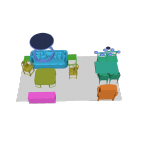}
    \includegraphics[width=\textwidth]{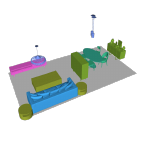}
    \vspace{-6pt}
    \caption{\centering DiffuScene~\cite{diffuscene}}
  \end{subfigure}
  \hfill
  \vrule
  \hfill
  \begin{subfigure}[t]{0.19\textwidth}
    \centering
    \includegraphics[width=\textwidth]{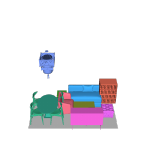}
    \includegraphics[width=\textwidth]{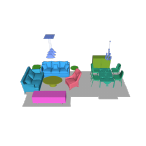}\vspace{1pt}
    \includegraphics[width=\textwidth]{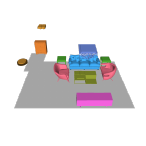}
    \includegraphics[width=\textwidth]{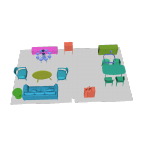}
    \includegraphics[width=\textwidth]{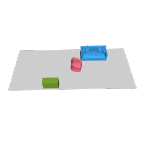}
    \includegraphics[width=\textwidth]{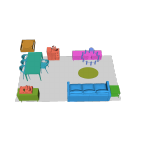}
    \includegraphics[width=\textwidth]{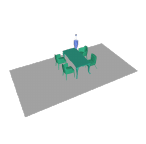}\vspace{2pt}
    \caption{\centering\textbf{DeBaRA}\protect\linebreak\textit{Dataset Random}}
  \end{subfigure}
  \hfill
  \vrule
  \hfill
  \begin{subfigure}[t]{0.19\textwidth}
    \centering
    \includegraphics[width=\textwidth]{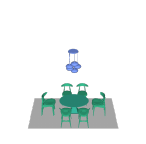}
    \includegraphics[width=\textwidth]{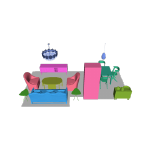}
    \includegraphics[width=\textwidth]{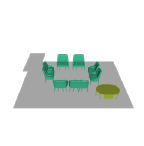}
    \includegraphics[width=\textwidth]{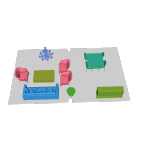}\vspace{-2pt}
    \includegraphics[width=\textwidth]{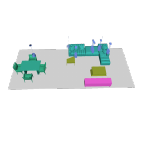}\vspace{-2pt}
    \includegraphics[width=\textwidth]{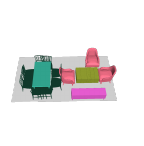}
    \includegraphics[width=\textwidth]{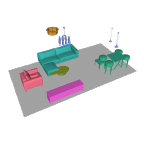}
    \vspace{-10pt}
    \caption{\centering\textbf{DeBaRA}\protect\linebreak\textit{LayoutGPT}}
  \end{subfigure}
  \hfill
  \vrule
  \hfill
  \begin{subfigure}[t]{0.19\textwidth}
    \centering
    \includegraphics[width=\textwidth]{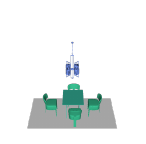}
    \includegraphics[width=\textwidth]{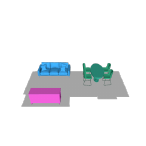}
    \includegraphics[width=\textwidth]{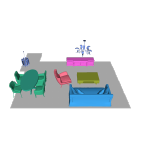}
    \includegraphics[width=\textwidth]{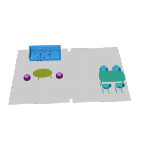}
    \includegraphics[width=\textwidth]{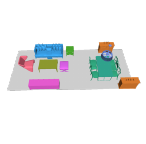}
    \includegraphics[width=\textwidth]{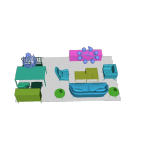}
    \includegraphics[width=\textwidth]{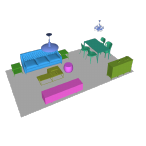}
    \vspace{-10pt}
    \caption{\centering\textbf{DeBaRA}\protect\linebreak\textit{SSE}}
  \end{subfigure}
  \vspace{4pt}
  \caption{\textbf{Additional scene synthesis results.} We compare DeBaRA with state-of-the-art approaches in various settings. Since object semantics are not part of our output space, they are randomly drawn from the training dataset (\textit{Dataset Random}), generated by an external LLM (\textit{LayoutGPT}) or selected from LLM-generated sets by our density estimate procedure (\textit{SSE}), which helps to get more natural layouts for the considered floor configuration.}
\end{figure}
\clearpage

\begin{figure}[H]
  \centering
  \begin{subfigure}[b]{0.26\textwidth}
    \centering
    \includegraphics[width=\textwidth]{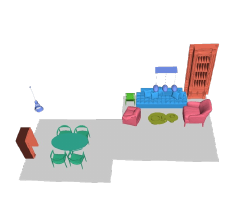}
    \includegraphics[width=\textwidth]{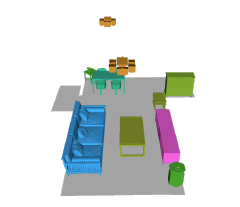}
    \includegraphics[width=\textwidth]{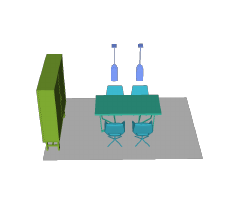}
    \caption{Ground Truth}
  \end{subfigure}
  \hfill
  \vrule
  \hfill
  \begin{subfigure}[b]{0.26\textwidth}
    \centering
    \includegraphics[width=\textwidth]{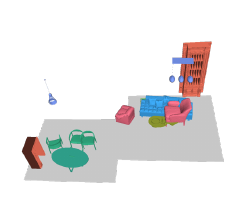}
    \includegraphics[width=\textwidth]{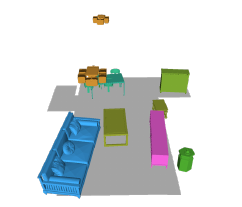}
    \includegraphics[width=\textwidth]{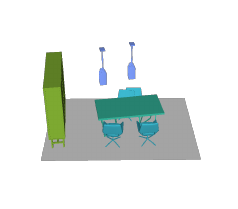}
    \caption{Perturbed}
  \end{subfigure}
  \hfill
  \vrule
  \hfill
  \begin{subfigure}[b]{0.26\textwidth}
    \centering
    \includegraphics[width=\textwidth]{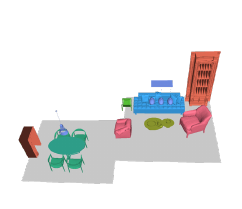}
    \includegraphics[width=\textwidth]{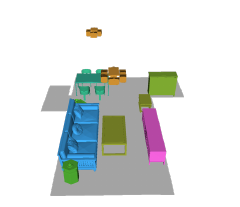}
    \includegraphics[width=\textwidth]{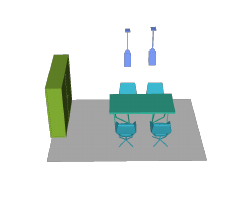}
    \caption{\textbf{Re-arranged}}
  \end{subfigure}
  \vspace{4pt}
  \caption{\textbf{Additional re-arrangement results.} Our method can recover a plausible arrangement from a noisy one using iterative sampling with fixed dimension attributes and low initial noise level.}  
\end{figure}

\begin{figure}[H]
  \centering
  \begin{minipage}{0.05\textwidth}
    \centering
    \rotatebox{90}{\small Partial}
  \end{minipage}
  \hfill
  \begin{minipage}{0.18\textwidth}
    \centering
    \includegraphics[width=\textwidth]{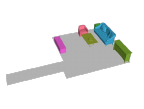}
  \end{minipage}
  \hfill
  \begin{minipage}{0.18\textwidth}
    \centering
    \includegraphics[width=\textwidth]{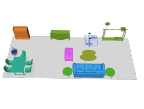}
  \end{minipage}
  \hfill
  \begin{minipage}{0.18\textwidth}
    \centering
    \includegraphics[width=\textwidth]{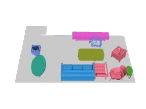}
  \end{minipage}
  \hfill
  \begin{minipage}{0.18\textwidth}
    \centering
    \includegraphics[width=\textwidth]{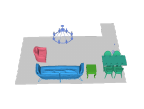}
  \end{minipage}
  \hfill
  \begin{minipage}{0.18\textwidth}
    \centering
    \includegraphics[width=\textwidth]{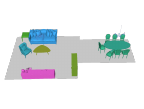}
  \end{minipage}
  \vspace{4pt}
  \hrule
  \vspace{4pt}
  \begin{minipage}{0.05\textwidth}
    \centering
    \rotatebox{90}{\small\textbf{Completed}}
  \end{minipage}
  \hfill
  \begin{minipage}{0.18\textwidth}
    \centering
    \includegraphics[width=\textwidth]{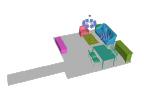}
  \end{minipage}
  \hfill
  \begin{minipage}{0.18\textwidth}
    \centering
    \includegraphics[width=\textwidth]{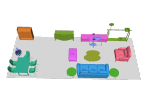}
  \end{minipage}
  \hfill
  \begin{minipage}{0.18\textwidth}
    \centering
    \includegraphics[width=\textwidth]{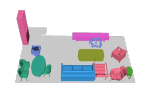}
  \end{minipage}
  \hfill
  \begin{minipage}{0.18\textwidth}
    \centering
    \includegraphics[width=\textwidth]{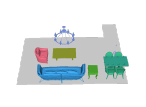}
  \end{minipage}
  \hfill
  \begin{minipage}{0.18\textwidth}
    \centering
    \includegraphics[width=\textwidth]{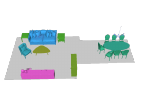}
  \end{minipage}
  \vspace{4pt}
  \caption{\textbf{Additional scene completion results.} From a list of additional objects semantics, DeBaRA is able to finely introduce the relevant items into an existing layout.}
  \vspace{4pt}
\end{figure}

\subsection{Statistical Analysis}

We can observe in our qualitative results, e.g., in Figure~\ref{fig:arrangegen}, that objects are most of the time parallel or perpendicular to walls, meaning that their rotation attributes \(\rr\) mainly take \textit{cardinal} angle values, i.e., \(0\)\textdegree, \(90\)\textdegree, \(180\)\textdegree{} or \(270\)\textdegree.

\begin{figure}[H]
  \centering
  \begin{subfigure}[b]{0.48\textwidth}
      \centering
      \caption*{\textbf{Living Rooms}}
      \begin{subfigure}[b]{0.48\textwidth}
          \centering
          \includegraphics[width=\textwidth]{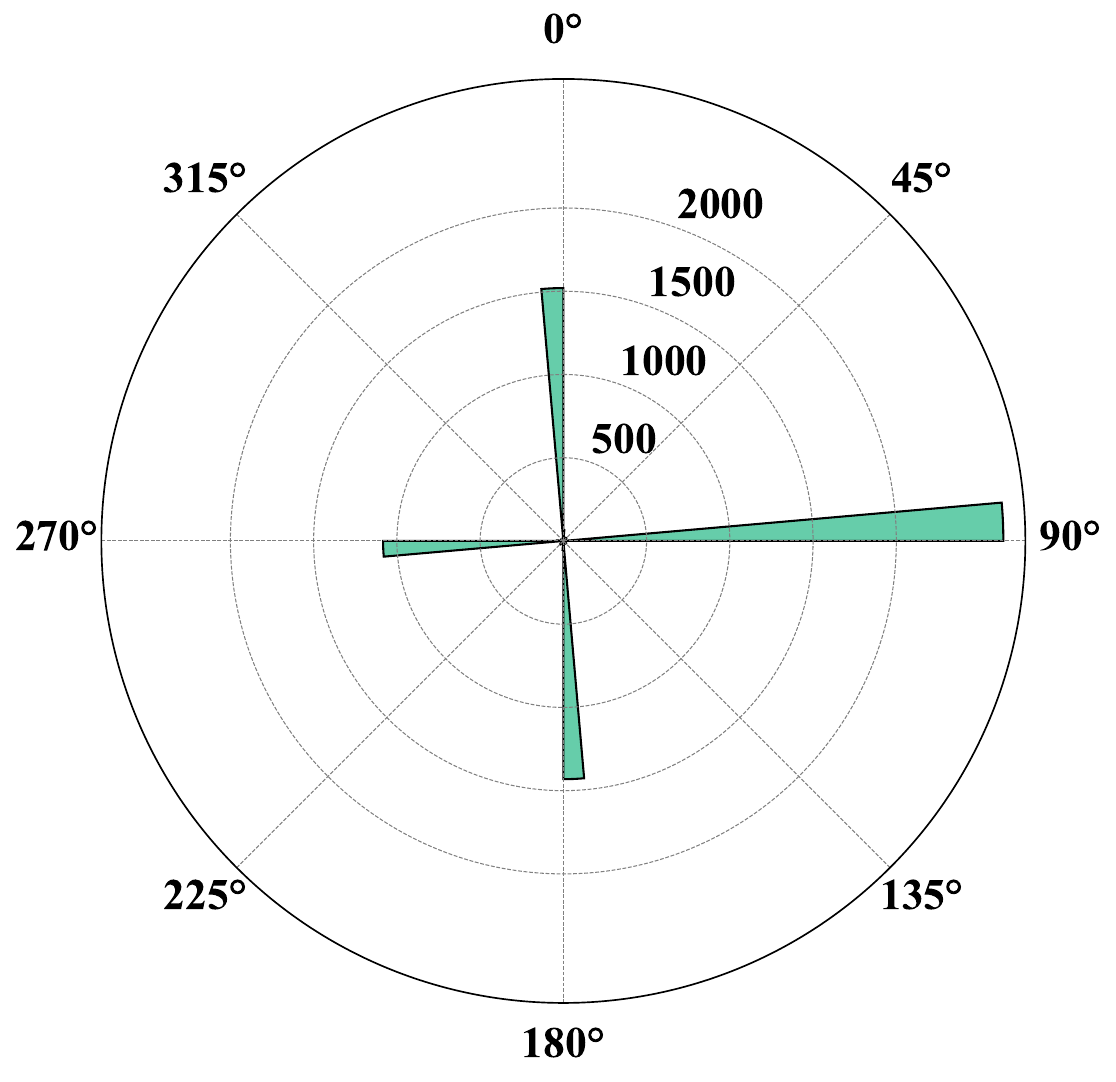}
          \caption{Dataset}
          \label{fig:exp1_left}
      \end{subfigure}
      \hfill
      \begin{subfigure}[b]{0.48\textwidth}
          \centering
          \includegraphics[width=\textwidth]{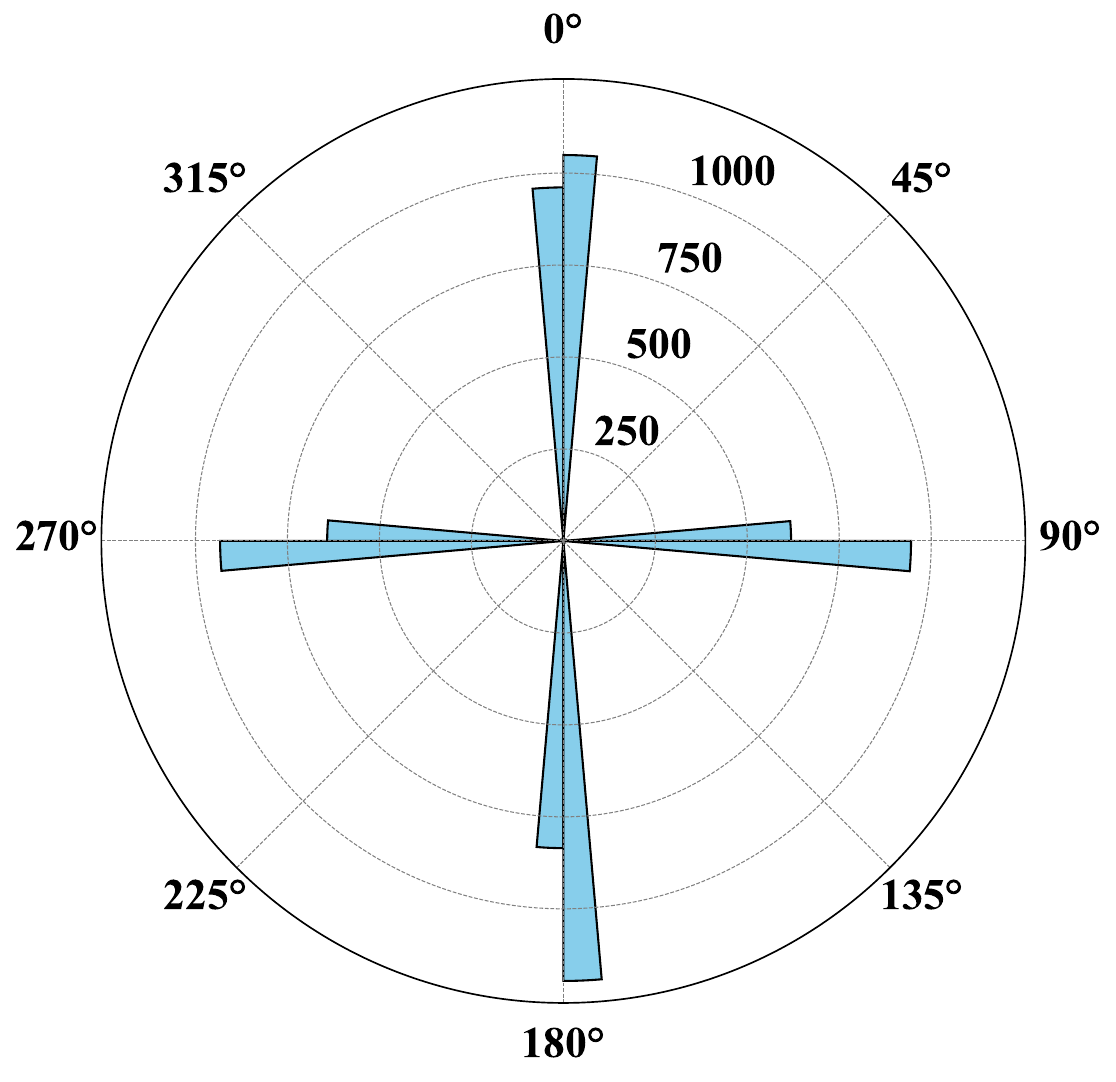}
          \caption{DeBaRA}
          \label{fig:exp1_right}
      \end{subfigure}
  \end{subfigure}
  \hfill
  \begin{subfigure}[b]{0.48\textwidth}
      \centering
      \caption*{\textbf{Dining Rooms}}
      \begin{subfigure}[b]{0.48\textwidth}
          \centering
          \includegraphics[width=\textwidth]{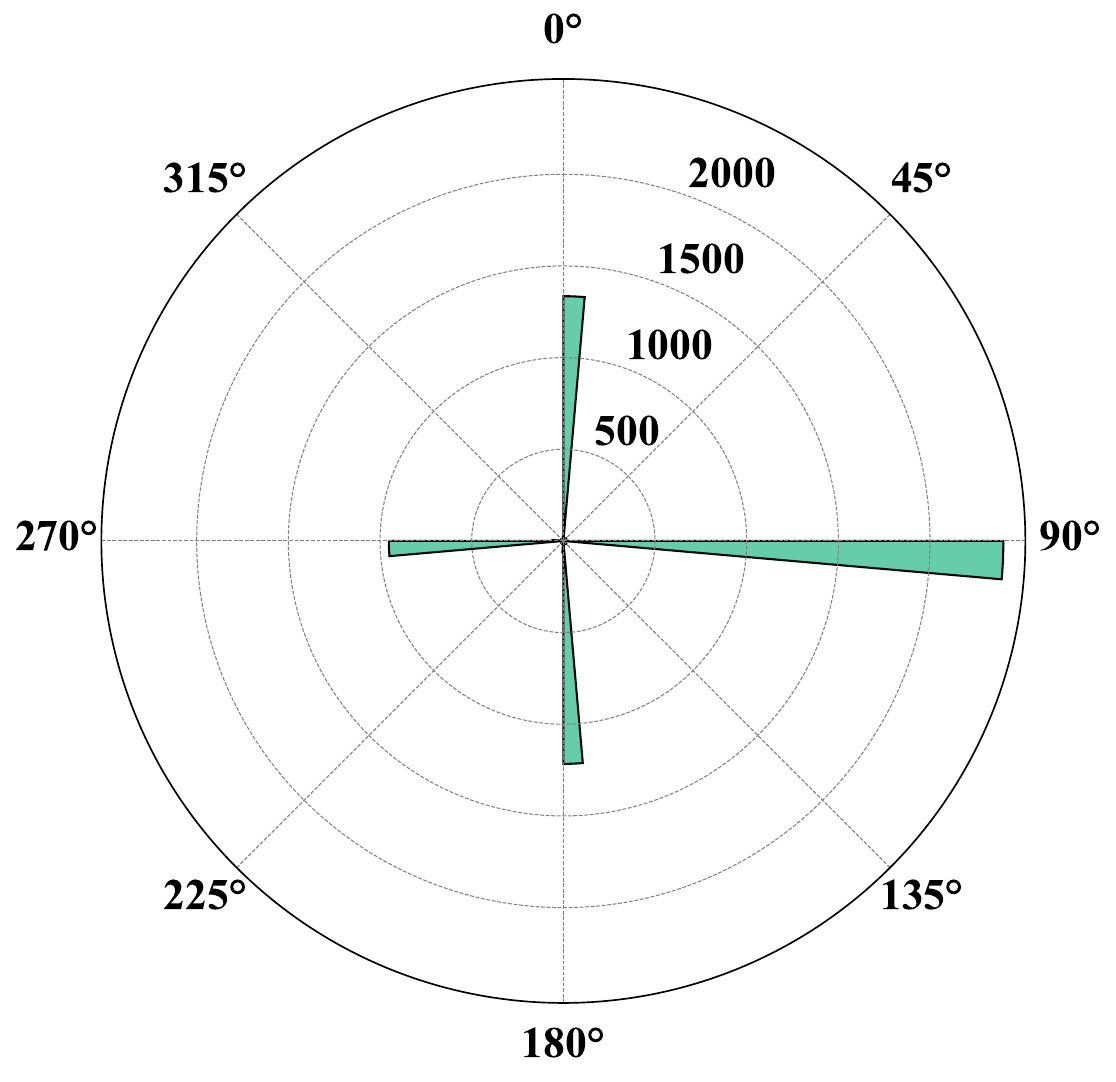}
          \caption{Dataset}
          \label{fig:exp2_left}
      \end{subfigure}
      \hfill
      \begin{subfigure}[b]{0.48\textwidth}
          \centering
          \includegraphics[width=\textwidth]{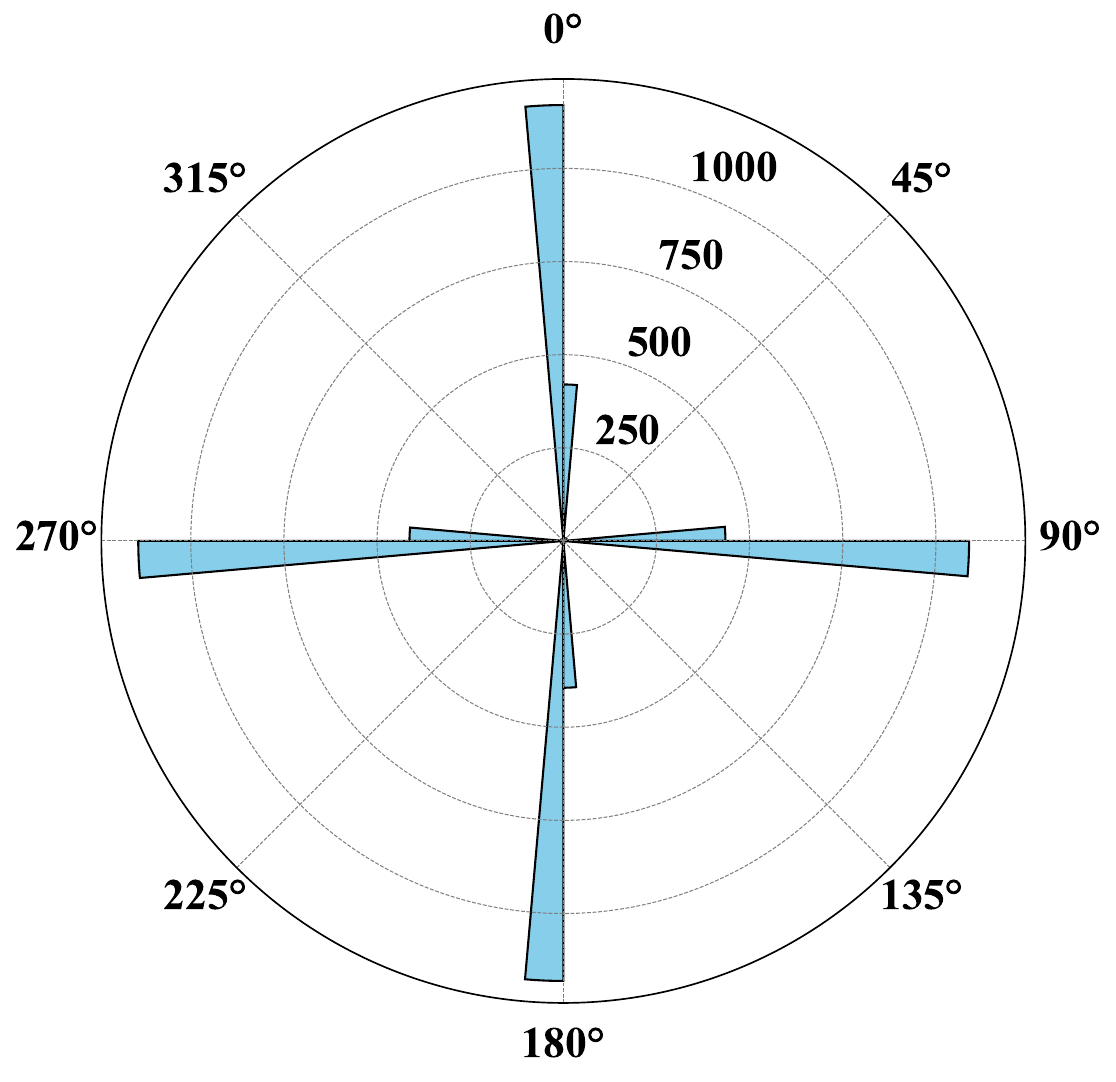}
          \caption{DeBaRA}
          \label{fig:exp2_right}
      \end{subfigure}
  \end{subfigure}
  \caption{Distributions of object rotation values extracted from the dataset or generated following our method, reported for both test subsets. We can see that the original data predominantly features objects rotated by \(0\)\textdegree, \(90\)\textdegree, \(180\)\textdegree{} or \(270\)\textdegree{} around the scene's vertical axis, which, as a result, is also the tendency induced by the generative model.}
  \label{fig:hist_ang}
\end{figure}

Radial histograms of object rotation values reported in Figure~\ref{fig:hist_ang} indicate that 3D-FRONT~\cite{3dfront} subsets contain rooms in which objects, in their vast majority, exhibit this property and don't have much \textit{exotic} angles, i.e., different from cardinal ones. These restricted rotation distributions are consequently learned and resembled by generative models at sampling time.
\clearpage

\section{Societal Impact}
\label{sec:sup:societal}

We believe that our method will predominantly yield positive societal impact by enhancing accessibility, controllability and efficiency in 3D indoor design and creation, benefiting both non-experts and professionals in several industries.

Nevertheless, it raises common concerns associated with deep generative models. Specifically, users should be warned about using the model for furnishing real-world rooms as generated layouts may result in unsafe designs. Additionally, privacy safeguards should systematically be implemented along our method in order to prevent unauthorized replication of personal spaces.

\section{Licenses}
\label{sec:sup:licenses}

In this section, we list licenses of datasets, open-source code artifacts and pretrained models that were used in the context of our experiments.

\paragraph{Datasets}
\begin{itemize}
  \item \makebox[36mm][l]{3D-FRONT~\cite{3dfront}:}CC BY-NC-SA 4.0 License
  \item \makebox[36mm][l]{3D-FUTURE~\cite{3dfuture}:}CC BY-NC-SA 4.0 License
\end{itemize}

\paragraph{Code}
\begin{itemize}
  \item \makebox[36mm][l]{ATISS~\cite{atiss}:}NVIDIA Source Code License for ATISS\footnote{\url{https://github.com/nv-tlabs/ATISS/blob/master/LICENSE}}
  \item \makebox[36mm][l]{DiffuScene~\cite{diffuscene}:}Sony Group Corporation License for DiffuScene\footnote{\url{https://github.com/tangjiapeng/DiffuScene/blob/master/LICENSE}}
  \item \makebox[36mm][l]{LayoutGPT~\cite{layoutgpt}:}MIT License
  \item \makebox[36mm][l]{LEGO-Net~\cite{legonet}:}MIT License
  \item \makebox[36mm][l]{PointNet~\cite{qi2017pointnet}:}MIT License
\end{itemize}

\paragraph{Models}
\begin{itemize}
  \item \makebox[36mm][l]{Llama-3-8B~\cite{llama3}:}Meta Llama 3 Community License Agreement\footnote{\url{https://huggingface.co/meta-llama/Meta-Llama-3-8B/blob/main/LICENSE}}
  \item \makebox[36mm][l]{Inception-v3~\cite{inception}:}Apache 2.0 License
\end{itemize}

\end{document}